\documentclass[twocolumn,journal]{IEEEtran}
\usepackage[T1]{fontenc}
\usepackage[latin9]{inputenc}
\usepackage{amsmath}
\usepackage{amssymb}
\usepackage{graphicx}

\usepackage[unicode=true,
 bookmarks=true,bookmarksnumbered=true,bookmarksopen=true,bookmarksopenlevel=1,
 breaklinks=false,pdfborder={0 0 0},pdfborderstyle={},backref=false,colorlinks=false]
 {hyperref}
\hypersetup{pdftitle={Your Title},
 pdfauthor={Your Name},
 pdfpagelayout=OneColumn, pdfnewwindow=true, pdfstartview=XYZ, plainpages=false}

\makeatletter
\usepackage[caption=false,font=footnotesize]{subfig}
\usepackage{booktabs}
\usepackage{multirow}
\usepackage{bbm} 
\usepackage[linesnumbered, ruled]{algorithm2e}
\usepackage{color}
\usepackage{xcolor}
\newcommand{\PyComment}[1]{\texttt{\textbf{\textcolor{teal}{{\# }#1}}}}  
\newcommand{\PyCode}[1]{\texttt{\textbf{\textcolor{black}{{}#1}}}} 
\usepackage{pifont}
\newcommand{\cmark}{\ding{51}}%
\newcommand{\xmark}{\ding{55}}%

\@ifundefined{showcaptionsetup}{}{%
 \PassOptionsToPackage{caption=false}{subfig}}
\usepackage{subfig}
\makeatother

\begin{document}
\title{Large-Scale Hyperspectral Image Clustering Using Contrastive Learning}
\author{Yaoming Cai,~\IEEEmembership{Student Member, IEEE},~Zijia Zhang,~Yan
Liu,~Pedram Ghamisi, \IEEEmembership{Senior Member, IEEE}, Kun Li,
Xiaobo Liu, \IEEEmembership{Member, IEEE}, and Zhihua Cai\thanks{This work was supported in part by the National Natural Science Foundation
of China (NSFC) under Grant 61773355 and Grant 61973285 and in part
by the National Scholarship for Building High Level Universities,
China Scholarship Council (No. 202006410044). (\emph{Corresponding
author: Zhihua Cai.})}\thanks{Yaoming Cai, Zijia Zhang, Yan Liu, and Zhihua Cai are with the School
of Computer Science, China University of Geosciences, Wuhan 430074,
China, and Yaoming Cai and Zijia Zhang are also with the Helmholtz-Zentrum
Dresden-Rossendorf (HZDR), Helmholtz Institute Freiberg for Resource
Technology, 09599 Freiberg, Germany (e-mail: caiyaom@cug.edu.cn; zhangzijia@cug.edu.cn;
yanliu@cug.edu.cn; zhcai@cug.edu.cn)}\thanks{Pedram Ghamisi is with the Helmholtz-Zentrum Dresden-Rossendorf (HZDR),
Helmholtz Institute Freiberg for Resource Technology, 09599 Freiberg,
Germany, and also with the Institute of Advanced Research in Artificial
Intelligence (IARAI), 1030 Vienna, Austria (e-mail: p.ghamisi@gmail.com).}\thanks{Kun Li is with the College of Electronic Science and Technology, National
University of Defense Technology, Changsha 410073, China. (e-mail:
likun19@nudt.edu.cn). }\thanks{Xiaobo Liu is with the School of Automation, China University of Geosciences,
Wuhan 430074, China, and also with the Hubei Key Laboratory of Advanced
Control and Intelligent Automation for Complex Systems, China University
of Geosciences, Wuhan 430074, China (e-mail: xbliu@cug.edu.cn).}}
\maketitle
\begin{abstract}
Clustering of hyperspectral images is a fundamental but challenging task. The recent development of hyperspectral image clustering has evolved from shallow models to deep and achieved promising results in many benchmark datasets. However, their poor scalability, robustness, and generalization ability, mainly resulting from their offline clustering scenarios, greatly limit their application to large-scale hyperspectral data. To circumvent these problems, we present a scalable deep online clustering model, named Spectral-Spatial Contrastive Clustering (SSCC), based on self-supervised learning. Specifically, we exploit a symmetric twin neural network comprised of a projection head with a dimensionality of the cluster number to conduct dual contrastive learning from a spectral-spatial augmentation pool. We define the objective function by implicitly encouraging within-cluster similarity and reducing between-cluster redundancy. The resulting approach is trained in an end-to-end fashion by batch-wise optimization, making it robust in large-scale data and resulting in good generalization ability for unseen data. Extensive experiments on three hyperspectral image benchmarks demonstrate the effectiveness of our approach and show that we advance the state-of-the-art approaches by large margins.
\end{abstract}


\section{Introduction}

\IEEEPARstart{R}{ecent} advances in the Earth observation (EO) technology
have provided end users with a large volume of remote sensing data
comprised of rich spectral, spatial, and temporal information \cite{HSIC-Review-Pedram-GRSM-17,HSIC-Review-Li-TGRS-19}.
As one of the most important technologies in EO, hyperspectral imaging
has facilitated numerous promising applications in areas ranging from
ocean, urban, agriculture, geology \cite{HSI-App-Review-IPT-20},
to biomedicine \cite{HSI-App-Bio-TMI-21}, due to its unique advantages.
Hyperspectral imagery (HSI) is characterized by hundreds of narrow
spectral bands with a nanometer resolution, allowing more fine-grained
recognition for objects of interest. 

HSI intelligent interpretation has drawn increasing attention in recent
years, wherein pixel-wise HSI classification lays a foundation for
decision-making. Parallel to the development of supervised learning,
supervised HSI classification has been extensively studied \cite{HSI-Review-Pedram-GRSM-19}
over the last decade. Despite the observable advances, supervised
methods commonly rely on sufficient training data. In particular,
with the growing success of deep learning in HSI processing \cite{HSI-Review-Nicolas-GRSM-19,HSI-BSNets-CAI-TGRS-20},
there is an increasing demand for large-scale training data paired
with human annotations \cite{HSI-LimitedSam-Jia-NEUC-21}. However,
data annotation often requires considerable manpower and resources,
as well as expert knowledge. This has become a severe bottleneck that
restricts the development of HSI intelligent interpretation. 

In contrast, there is a distinct gap between HSI clustering (i.e.,
unsupervised HSI classification) and supervised HSI classification
\cite{HSIClu-GCSC-CAI-TGRS-20}. Due to the absence of supervisory
signals and also the ubiquitous spectral variability \cite{HSI-SpectralVariability-GRSM-21},
HSI clustering is much more challenging than supervised tasks. The core of HSI clustering is
to model the intrinsic interrelation between data points and clusters.
Therefore, HSI clustering often involves pair-wise measurement criteria.
For example, classic $k$-means \cite{kmeans-TPAMI-02} and fuzzy
$c$-means \cite{FCM-ICFS-09} seek to find a set of cluster centers
by calculating pair-wise distance. Instead, subspace clustering, e.g.,
sparse subspace clustering (SSC) \cite{SSC-Ehsan-TPAMI-13} and low-rank
subspace clustering (LRSC) \cite{LRSC-PRL-14}, aims to learn the
self-expressiveness of data. Beneficial from the good theoretical guarantees
and robustness, subspace clustering has been widely applied for HSI
clustering \cite{HSIClu-GCSC-CAI-TGRS-20,HSIClu-S4C-ZHang-TGRS-16,HSIClu-JSSC-ZHai-TGRS-21}.
Despite their promising results, most of HSI clustering methods typically
suffer from two drawbacks.

First, previous methods usually model in either raw feature space
or hand-crafted feature space. This would produce inferior results
on complex and high-dimensional HSI datasets caused by insufficient
representability. To overcome the limitation, many efforts \cite{HSIClu-GR-RSCNet-CAI-INS-21,HSIClu-DSC-Lei-TCSVT-21,HSIClu-SSL-DSC-JSTARS-21,HSIClu-DCIDC-Sun-TGRS-21}
have been devoted to developing deep clustering models. The common
idea is to incorporate a clustering objective function, often derived
from classical clustering mechanisms with a proper relaxation, into
a deep learning model and then jointly train for discriminative representations
\cite{DeepClu-Review-18,DeepClu-Review-ACC-18}. For instance, deep
subspace clustering networks (DSC) \cite{DSC-Pan-NIPS-17} and its
variants \cite{DSC-Peng-TNNLS-20,HSIClu-GCSC-CAI-TGRS-20,DSC-PseudoSu-TIP-21}
recast subspace learning into a latent self-expression layer of an
autoencoder (AE). Similarly, soft and regularized $k$-means objective
\cite{DeepKmeans-Jabi-TPAMI-19} is frequently adopted in learning
clustering-friendly discriminative representations. This results in
a series of state-of-the-art deep clustering models including deep
embedded clustering (DEC) \cite{DeepClu-DEC-Xie-ICML-16} and deep
clustering network (DCN) \cite{DCN-Yang-ICML-17}. However, the training
schedule alternating between feature clustering and network parameters
update leads to unstable learning of feature representations \cite{DeepClu-ODC-Zhan-CVPR-20,SSL-CC-Li-AAAI-21}.

Second, most of the existing HSI clustering methods belong to offline
clustering models, resulting in two inevitable issues: poor scalability
and generalization ability. However, these two properties are crucial
for large-scale HSI clustering and application in industrial scenarios.
Many attempts have been dedicated to addressing these limitations.
For scalability, they presented some useful solutions from either
data perspective, e.g., dimensionality reduction \cite{HSIClu-RNMF-Zhang-INS-19,HSIClu-NMFAML-Qin-JSTARS-21}
and superpixel segmentation \cite{HSIClu-GCSSC-Huang-TGRS-21}, or
model mechanism perspective including anchor graph \cite{HSIClu-SGCNR-Wang-TGRS-19}
and sampling \cite{Subspace-S5C-NIPS-19,Subspace-ESC-ECCV18,IDEC-IJCAI-17}.
Whereas, there are fewer studies that have focused on the generalization
ability of clustering models, especially in the HSI area. This is
because most clustering methods can only handle offline tasks, i.e.,
the clustering is based on the whole dataset, leading to difficulty
in dealing with unseen data. Moreover, pure unsupervised training
for clustering may produce trivial solutions.

Recently, self-supervised learning \cite{SSL-Survey-Jing-TPAMI-21}
typified by contrastive learning \cite{SSL-SimCLR-Chen-ICML-20,SSL-BarlowTwins-Jure-ICML-21}
has emerged as a powerful paradigm for representation learning. It
also opens up a new path to smoothly overcome the aforementioned shortcomings.
Based on the philosophy of ``label is representation'' \cite{SSL-CC-Li-AAAI-21,SSL-COMPLETER-Lin-CVPR-21,SSL-DSEC-Chang-TPAMI-20},
the pioneer of using contrastive learning for clustering, i.e., contrastive
clustering (CC) \cite{SSL-CC-Li-AAAI-21}, has shown state-of-the-art
results in image clustering tasks. The approach conducts clustering
through both instance-level and cluster-level contrastive learning,
thereby avoiding explicit clustering objectives and resulting in remarkable
flexibility on training and applying. 

In this paper, we develop a one-stage online deep clustering approach for large-scale HSIs, termed as Spectral-Spatial Contrastive Clustering (SSCC), which recasts the clustering
task into a self-supervised learning problem. The intuition behind
the clustering model is based on an observation that a neural network
always tends to activate the same neurons if the patterns belong to
the same clusters. This signifies that the outputs of a softmax projection
head with a dimensionality of the cluster number can naturally treat
as soft pseudo labels. Following with this observation and
the pioneer CC \cite{SSL-CC-Li-AAAI-21} approach, we present a new
contrastive clustering framework based on exploiting twin neural networks
to implicitly maximize within-cluster similarity and reduce between-cluster
redundancy from a batch of distorted views augmented according to
spectral and spatial properties. Unlike the CC approach that adopts
dual InfoNCE \cite{InfoNCE-arXiv-18} objectives and two asymmetric
projection heads, our model is entirely symmetric containing only
one projection head and includes a different objective based on the
redundancy reduction suggested in Barlow Twins \cite{SSL-BarlowTwins-Jure-ICML-21}.
This alleviates the reliance on large batches and the risk of collapsed
representations. 

To summarize, our contributions are threefold:
\begin{itemize}
\item We propose SSCC approach for large-scale HSI clustering by using a self-supervised learning framework to learn
discriminative label representations. The approach can be easily trained
in an entirely end-to-end fashion by batch-wise optimization, making
it scalable for large-scale HSIs and easy to generalize to unseen data.
Instead of using dual projection heads and asymmetrical networks,
we show that a single projection head with a symmetry architecture
can produce better results. 
\item We introduce a novel objective function by leveraging both InfoNCE
and Barlow Twins objectives, which implicitly encourage within-cluster
similarity and between-cluster dissimilarity, respectively. To fully
explore spectral and spatial information, we provide a semantic-preserving
augmentation pool by distorting along both spectral and spatial dimensions. 
\item We systematically evaluate our approach on three HSI benchmarks, demonstrating
that SSCC consistently outperforms the state-of-the-art approaches
by large margins. We also show that SSCC indeed learns fine-grained
semantic differences from our designed self-supervised task. The successful
attempt provides an effective tool to narrow the gap between supervised
and unsupervised learning.
\end{itemize}
The rest of this paper is structured as follows. First, we briefly
review the related concepts, e.g., contrastive learning and HSI clustering
methods, in Section \ref{sec:Related-Works}. Second, we describe
the details of the developed SSCC approach in Section \ref{sec:Spectral-Spatial-Contrastive-Clu}.
In Section \ref{sec:Results}, we provide systematic experimental
results and analysis for our approach. Finally, we conclude with a
summary and future works in Section \ref{sec:Conclusions}.

\begin{figure*}[tbh]
\begin{centering}
\includegraphics[width=1.8\columnwidth]{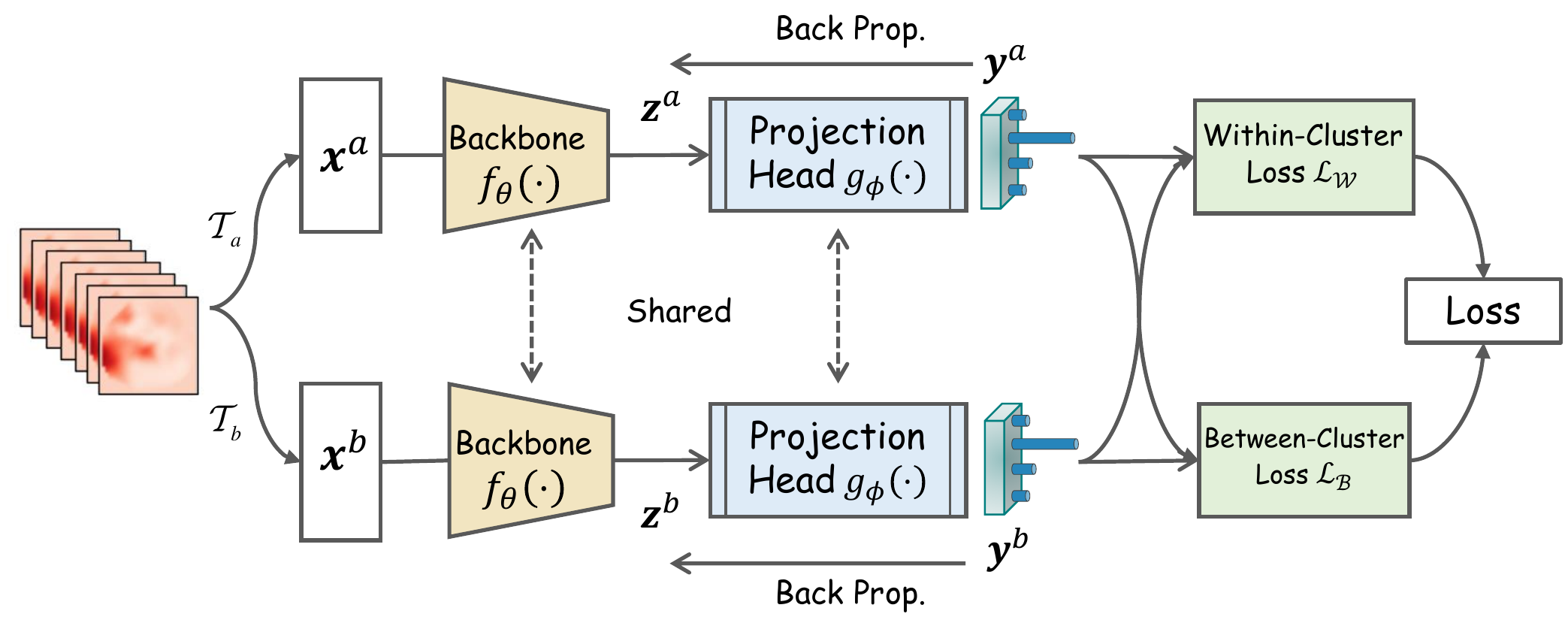}
\par\end{centering}
\caption{Overview of SSCC. Given a HSI cell $\boldsymbol{x}$ sampled
from an HSI, we feed it into an augmentation pool $\mathcal{T}$ twice
to generate two distorted views $\boldsymbol{x}^{a}$ and $\boldsymbol{x}^{b}$.
Then, we forward them into a network twin consisting of a backbone
$f_{\theta}\left(\cdot\right)$ followed by a projection head $g_{\phi}\left(\cdot\right)$
with $C$ softmax output neurons to obtain a pair of corresponding
latent representations (i.e., $\boldsymbol{z}^{a}$ and $\boldsymbol{z}^{b}$)
and label representations (i.e., $\boldsymbol{y}^{a}$ and $\boldsymbol{y}^{b}$).
SSCC is trained in an end-to-end manner by optimizing an objective
that uses $\mathcal{L}_{\mathcal{W}}$ to maximize within-cluster similarity
and $\mathcal{L}_{\mathcal{B}}$ to minimize between-cluster correlation.
During the inference stage, SSCC assigns label by taking the maximum activation
of the label representations, making it robust to large-scale data
and unseen data. \label{fig:Overview-of-SSCC}}
\end{figure*}

\section{Related Work \label{sec:Related-Works}}

\subsection{Self-supervised Contrastive Learning}

Self-supervised learning \cite{SSL-Survey-Jing-TPAMI-21}, as a subset
of unsupervised learning, has emerged as a promising paradigm for
representation learning. The goal of the paradigm is to learn general
feature representation from large-scale unlabeled data without using
any human-annotated labels. The pipeline of self-supervised learning
can be broadly summarized into two steps. First, it trains a neural
network to solve a predefined pretext task, such as predicting image
rotations \cite{SSL-Rotation-ICLR-18} and clustering labels
\cite{SSL-DeepCluster-ECCV-18}. Second, the pre-trained model is
transferred to a specific downstream task, e.g., classification, segmentation,
and object detection \cite{SSL-SimCLR-Chen-ICML-20,SSL-BarlowTwins-Jure-ICML-21,SSL-Rotation-ICLR-18},
by fine-tuning. On the basis of objectives, self-supervised learning
can summarize into three categories \cite{SSL-Review-TKDE-21}: generative,
contrastive, and adversarial. 

Due to the remarkable representation ability, self-supervised contrastive
learning has attracted increasing interest \cite{SSL-Review-TKDE-21}.
Typically, the method involves data augmentations and twin network
architectures (e.g., SimCLR \cite{SSL-SimCLR-Chen-ICML-20}), while
its objective frequently adopts InfoNCE \cite{InfoNCE-arXiv-18} defined
upon positive and negative pairs. Consequently, contrastive learning
usually requires sufficient batch samples to prevent trivial solutions.
To address the problem, momentum contrast (MoCo) suggests using a memory
bank and momentum update mechanism. Recent efforts including bootstrap
your own latent (BYOL) \cite{SSL-BYOL-NIPS-20} and simple siamese
(SimSiam) \cite{SSL-SimSiam-CVPR-21} show that asymmetric architecture
and a mechanism called 'stop-gradient' are probably the key factors
of avoiding negative samples and collapsed representations. In Barlow
Twins \cite{SSL-BarlowTwins-Jure-ICML-21} approach, Jure \emph{et
al. }further simplified the training of contrastive learning by a symmetric
architecture and an innovative redundancy reduction loss function. 

Increasing evidence suggests that self-supervised learning has tremendous
potential to break the dependencies on human-annotated labels. In the
HSI area, self-supervised learning has achieved promising results
across imaging \cite{HSI-Snapshot-SSL-ICCV-21}, unmixing \cite{HSI-Unmixing-SSL-TNNLS-21},
and classification \cite{HSIC-SSL-Distillation-TGRS-21}. These indicate
that the use of self-supervised learning for HSI clustering would make
great sense for narrowing the gap between unsupervised learning and
supervised learning of HSI. 

\subsection{Deep Clustering}

Unlike traditional clustering approaches, e.g., $k$-means \cite{kmeans-TPAMI-02},
deep clustering refers to the clustering model based on deep learning,
wherein deep learning is used to learn high-level features. Nonetheless,
traditional clustering models are usually reformulated as a clustering
regularization in the deep clustering \cite{DeepClu-Review-18}. A
typical deep clustering objective can be expressed as $\mathcal{L}=\mathcal{L}_{clu}+\gamma\mathcal{L}_{rep}$,
where $\mathcal{L}_{clu}$ and $\mathcal{L}_{rep}$ indicate a clustering
loss and a representation loss, respectively. Taking DSC \cite{DSC-Pan-NIPS-17}
as an example, $\mathcal{L}_{clu}$ is implemented by self-representation
in a latent space and $\mathcal{L}_{rep}$ is a reconstruction error
of AE. 

The training of deep clustering may involve two stages or one stage.
For the two-stage models like DeepCluster \cite{SSL-DeepCluster-ECCV-18},
they conduct network parameters update and feature clustering alternately.
Despite their flexibility, these models often suffer from poor stability
in clustering results and feature representation \cite{DeepClu-ODC-Zhan-CVPR-20}.
In contrast, one-stage models pursue clustering-friendly feature representation
by jointly feature learning and clustering, significantly improving
the clustering accuracy. Two representative instances of the one-stage
model are DEC \cite{DeepClu-DEC-Xie-ICML-16} and DSC \cite{DSC-Pan-NIPS-17},
which developed from $k$-means and subspace clustering, respectively.
The former proposes to learn cluster centroids in an embedding space
by minimizing Kullback\textendash Leibler divergence between soft
assignment and auxiliary target distribution, while the latter calculates
affinity matrix in a linear subspace. 

Recently, HSI clustering has evolved from shallow clustering to
deep clustering \cite{HSIClu-Review-ZHai-GSM-21}. In \cite{HSIClu-DSC-Lei-TCSVT-21,HSIClu-GR-RSCNet-CAI-INS-21,HSIClu-SSL-DSC-JSTARS-21},
various deep subspace clustering models are presented for HSI clustering. Despite remarkable clustering accuracy in small HSI scenes, these methods are commonly limited by their quadratic computation
and space complexity resulting from the full-batch self-representation.
Furthermore, most of the existing HSI clustering methods can only deal
with offline tasks, leading to poor robustness in large-scale HSI
data. These motivate us to develop a scalable and end-to-end
deep clustering model for large-scale HSI. 

\section{Proposed Method\label{sec:Spectral-Spatial-Contrastive-Clu}}

\subsection{Problem Definition}

HSI clustering is a process of grouping every HSI cell $\chi=\left\{ \boldsymbol{x}_{i}\in\mathbb{R}^{n_{1}\times n_{2}\times b}\right\} _{i=1}^{N}$
into $C$ distinct clusters such that patterns within the same clusters
are similar to each other, while those in different clusters are dissimilar,
where $N$ denotes the number of cells, each of which consists of
$n_{1}\times n_{2}$ pixels and $b$ spectral bands. Therefore, the
core of HSI clustering is to model for distinguishing within-cluster
similarity and between-cluster dissimilarity. From this perspective,
SSCC recasts the clustering task into a contrastive learning framework
which follows the observation of ``label is representation'' \cite{SSL-CC-Li-AAAI-21}.
The schematic of SSCC is illustrated in Fig. \ref{fig:Overview-of-SSCC}
and more details are introduced as follows. 

\subsection{Overall Framework}

The SSCC model follows a typical contrastive learning framework. Specifically,
it consists of a predefined augmentation pool $\mathcal{T}$ for distorting
original inputs, and a twin network composed of a backbone network
$f_{\theta}\left(\cdot\right)$ parameterized by $\theta$ followed
by a projection head $g_{\phi}\left(\cdot\right)$ parameterized $\phi$.
More precisely, let $\mathcal{T}^{a},\mathcal{T}^{b}\in\mathcal{T}$
be two augmentation compositions sampled from the augmentation pool
$\mathcal{T}$. By applying $\mathcal{T}^{a}$ and $\mathcal{T}^{b}$
to $\boldsymbol{x}$, we obtain two corresponding distorted views
of $\boldsymbol{x}$, denoted as $\boldsymbol{x}^{a}=\mathcal{T}^{a}\left(\boldsymbol{x}\right)$
and $\boldsymbol{x}^{b}=\mathcal{T}^{b}\left(\boldsymbol{x}\right)$.
Next, we forward them into the backbone network to capture deep spectral
and spatial semantic representations, resulting in latent representations
$\boldsymbol{z}^{a}=f_{\theta}\left(\boldsymbol{x}^{a}\right)$ and
$\boldsymbol{z}^{b}=f_{\theta}\left(\boldsymbol{x}^{b}\right)$. The
latent representations are subsequently fed into a special projection
head that is comprised  of $C$ output neurons activated by the softmax
function, defined as $g_{\phi}:\boldsymbol{z}\mapsto\boldsymbol{y}\in\mathbb{R}^{C}$,
to predict the corresponding label representations, i.e., $\boldsymbol{y}^{a}=g_{\phi}\left(\boldsymbol{z}^{a}\right)$
and $\boldsymbol{y}^{b}=g_{\phi}\left(\boldsymbol{z}^{b}\right)$.
We treat the label representations as soft pseudo labels of inputs,
wherein each output neuron associates with a certain land cover object.
The intuition behind this insight is that neural networks tend to
capture the fine-grained semantic differences between underlying objects.
This signifies that, although there is no explicit supervisory information,
input patterns belonging to the same clusters tend to hold the same
activated neurons. Hence, the key to SSCC is how to increase the discriminative
ability of label representations. 

\begin{algorithm}[htp!] 
	\SetAlgoLined  
	\PyComment{f: backbone network} \\ 
	\PyComment{g: projection head} \\      
	\PyComment{criterion\_wi: within-cluster loss} \\ 
	\PyComment{criterion\_bt: between-cluster loss} \\ 
	\PyComment{temp,lamb,alph: parameters in loss} \\
	\PyComment{M: batch size} \\
	\PyComment{}\\        
	\PyComment{training stage} \\ 
	\PyComment{load a batch with M samples}\\
	\PyCode{for x in loader:}\\  
	\Indp
	\PyComment{two randomly augmented views of x}\\
	\PyCode{x\_a, x\_b = transform(x)}\\
	\PyComment{forward data}\\
	\PyCode{y\_a = g(f(x)), y\_b = g(f(x))}\\
	\PyComment{calculate within-cluster loss}\\
	\PyCode{loss\_wi= criterion\_wi(y\_a,y\_b,temp)}\\
	\PyComment{calculate between-cluster loss}\\
	\PyCode{loss\_bt= criterion\_bt(y\_a,y\_b,lamb)}\\
	\PyComment{calculate final loss}\\
	\PyCode{loss = loss\_bt + alph * loss\_wi}\\
	\PyComment{optimization step}\\
	\PyCode{loss.backward()}\\
	\PyCode{optimizer.step()}\\
	\Indm 
	\PyComment{} \\  
	\PyComment{inference stage} \\   
	\PyCode{logit = g(f(x\_test))}\\ 
	\PyCode{labels = argmax(logit, dim=1)}\\
	\caption{PyTorch-style pseudocode for SSCC} 
	\label{algo:sscc} 
\end{algorithm}

\subsection{Objective Function}

Our objective function is designed to implicitly encourage within-cluster
similarity and between-cluster dissimilarity simultaneously for the
label representations. More specifically, we propose to use a within-cluster
contrastive objective $\mathcal{L}_{\mathcal{W}}$ and a between-cluster
contrastive objective $\mathcal{L}_{\mathcal{B}}$ to achieve this
purpose. The resulting objective function is defined upon the label
representations as the sum over these two contrastive terms, i.e.,
\begin{equation}
\mathcal{L}=\mathcal{L}_{\mathcal{B}}+\alpha\mathcal{L}_{\mathcal{W}},
\end{equation}
where $\alpha$ denotes a balance coefficient. A detailed introduction
is described as follows. \\

\subsubsection{Within-Cluster Contrastive Loss}
We adopt a typical contrastive loss called InfoNCE \cite{InfoNCE-arXiv-18,SSL-SimCLR-Chen-ICML-20}
to achieve $\mathcal{L}_{\mathcal{W}}$. Formally, given input HSI
cells of a batch of size of $M$, denoting as $\mathcal{D}=\left\{ \boldsymbol{x}_{i}\right\} _{i=1}^{M}$,
and its distorted version indicated as $\widetilde{\mathcal{\mathcal{D}}}=\left\{ \boldsymbol{x}_{1}^{a},...,\boldsymbol{x}_{M}^{a},\boldsymbol{x}_{1}^{b},...,\boldsymbol{x}_{M}^{b}\right\} $, Take arbitrary instance $\boldsymbol{x}_{i}^{k}\in\mathcal{\widetilde{\mathcal{\mathcal{D}}}},s.t.,k\in\left\{ a,b\right\} $
as anchor. We refer to the pair of $\left(\boldsymbol{x}_{i}^{k_{1}},\boldsymbol{x}_{i}^{k_{2}}\right),s.t.,k_{1}\neq k_{2}\in\left\{ a,b\right\} $
as a positive pair, while treating other $2(M-1)$ pairs as negative
pairs with regard to this positive pair. The goal of $\mathcal{L}_{\mathcal{W}}$
is to maximize the similarity between positive pairs. Specifically,
given a pair-wise similarity criterion for two vectors $\boldsymbol{u}$
and $\boldsymbol{v}$, e.g., the cosine similarity $s\left(\boldsymbol{u},\boldsymbol{v}\right)=\boldsymbol{u}\boldsymbol{v}^{T}/\left\Vert \boldsymbol{u}\right\Vert \left\Vert \boldsymbol{v}\right\Vert $,
$\mathcal{L}_{\mathcal{W}}$ for a positive pair $\left(\boldsymbol{x}_{i}^{a},\boldsymbol{x}_{i}^{b}\right)$
is formulated as{\scriptsize{}
\begin{equation}
\mathcal{\mathcal{L}}_{\mathcal{W}}^{a}\left(i\right)\!=\!-\log\left(\!\frac{\exp\left(\mathrm{s}\left(\boldsymbol{y}_{i}^{a},\!\boldsymbol{y}_{i}^{b}\right)/\tau\right)}{\sum_{j=1}^{M}\left[\mathbbm{1}_{j\neq i}\exp\left(\mathrm{s}\left(\boldsymbol{y}_{i}^{a},\!\boldsymbol{y}_{i}^{a}\right)/\tau\right)\!+\!\exp\left(\mathrm{s}\left(\boldsymbol{y}_{i}^{a},\boldsymbol{y}_{j}^{b}\right)/\tau\right)\right]}\!\right),
\end{equation}
}where $\mathbbm{1}_{j\neq i}=\left\{ 0,1\right\} $ is an indicator
function to 1 iff $j\neq i$ and $\tau$ indicates a temperature parameter
that controls the scale of distribution. Similarly, we use $\mathcal{\mathcal{L}}_{\mathcal{W}}^{b}\left(i\right)$
to indicate the contrastive loss for a positive pair $\left(\boldsymbol{x}_{i}^{b},\boldsymbol{x}_{i}^{a}\right)$.
Finally, the within-cluster contrastive loss is computed and averaged
across all positive pairs in a mini-batch, both $\left(a,b\right)$
and $\left(b,a\right)$, i.e.,
\begin{equation}
\mathcal{L}_{\mathcal{W}}=\sideset{\frac{1}{2M}}{_{i=1}^{M}}\sum\left[\mathcal{\mathcal{L}}_{\mathcal{W}}^{a}\left(i\right)+\mathcal{\mathcal{L}}_{\mathcal{W}}^{b}\left(i\right)\right].
\end{equation}
It is easy to prove that minimizing $\mathcal{L}_{\mathcal{W}}$ is
equivalent to implicitly encourage the consistency within the same
clusters.

\subsubsection{Between-Cluster Contrastive Loss}

The between-cluster contrastive loss $\mathcal{\mathcal{L}}_{\mathcal{B}}$
is designed to push away different clusters. There are various options
to achieve this, e.g., using InfoNCE again in column space as it was in CC \cite{SSL-CC-Li-AAAI-21}. However, InfoNCE has poor robustness
and stability with respect to the batch size. Here, instead, we borrow the decorrelation
mechanism suggested in Barlow Twins \cite{SSL-BarlowTwins-Jure-ICML-21}
to implement $\mathcal{\mathcal{L}}_{\mathcal{B}}$. Formally, we
rewrite the centered label representations for a mini-batch of samples
into matrix forms, i.e., $\mathbf{Y}^{k}\in\mathbb{R}^{M\times C},s.t.,k\in\left\{ a,b\right\} $.
Ideally, $\mathbf{Y}^{a}$ and $\mathbf{Y}^{b}$ should be identical
and tend to be binary matrices, where each column $\boldsymbol{y}_{\cdot j}^{k}$
can be treated as the marginal probability distribution of the batch
samples across the $j$-th cluster. Let $\mathcal{C}\in\mathbb{R}^{C\times C}$
be a cross-correlation matrix computed between $\mathbf{Y}^{a}$ and
$\mathbf{Y}^{b}$ along the batch dimension, where each entity $\mathcal{C}_{ij}$
associated with the $i$-th and the $j$-th cluster is computed as
\footnote{Note that the cross-correlation matrix for two observation variables
is equal to the pair-wise cosine similarity matrix for their centered
version. } 
\begin{equation}
\mathcal{C}_{ij}=\frac{\boldsymbol{y}_{\cdot i}^{a}\left(\boldsymbol{y}_{\cdot j}^{b}\right)^{\mathrm{T}}}{\left\Vert \boldsymbol{y}_{\cdot i}^{a}\right\Vert \left\Vert \boldsymbol{y}_{\cdot j}^{b}\right\Vert }.
\end{equation}
Based on the cross-correlation matrix, $\mathcal{\mathcal{L}}_{\mathcal{B}}$
is defined as the following decoupled form with a tradeoff coefficient
$\lambda$, i.e., 
\begin{equation}
\mathcal{\mathcal{L}}_{\mathcal{B}}=\sideset{}{_{i=1}^{M}}\sum\left(\mathcal{C}_{ii}-1\right)^{2}+\sideset{\sideset{\lambda}{_{i=1}^{M}}\sum}{_{j\neq i}^{M}}\sum\mathcal{C}_{ii}^{2}.
\end{equation}
Intuitively, $\mathcal{\mathcal{L}}_{\mathcal{B}}$ encourages the
cross-correlation matrix of label representations to be close to the
identity matrix. This implies that the correlation between the same
cluster should be close to $1$ (perfect correlation) while decorrelating
between different clusters. Therefore, $\mathcal{\mathcal{L}}_{\mathcal{B}}$
implicitly maximizes the dissimilarity between different clusters
by reducing redundancy. The optimum of $\mathcal{\mathcal{L}}_{\mathcal{B}}$
will be achieved when label representations approximate the ground truth.

\subsection{Clustering Using SSCC Model}

Given any number of HSI cells $\mathbf{X}$ acquired from the same
sensor, even if they are unseen by the model, we can easily carry
out online clustering using the optimized SSCC. The cluster labels
$\mathcal{Y}$  of $\mathbf{X}$ can be formally determined by 
\begin{equation}
\mathcal{Y}=\underset{c\in\left\{ 1,\cdots,C\right\} }{\arg\max}~g\left(f\left(\mathbf{X}\right)\right).
\end{equation}
Instead of modeling for a fixed dataset as most offline clustering
models, our SSCC model is quite flexible in both training and applying,
resulting in a good scalability and generalization ability.

\subsection{Implementation Details}

We show the PyTorch-style pseudocode for SSCC in Algorithm \ref{algo:sscc}.
More implementation details are described as follows and can also
be found from our release of the code\footnote{https://github.com/AngryCai/SSCC}. 

\subsubsection{Spectral-Spatial Augmentation}

Augmentation plays an important role in our SSCC model. Any transformation
that can distort original data but does not corrupt its semantic and
data shape is acceptable for augmentation. Thus, a good augmentation
often expands the diversity of the data distribution, thereby facilitating to extract semantic feature by twin networks. Based on this principle,
we design our augmentation pool $\mathcal{T}$ with both spatial and
spectral transformations by considering the characteristics of HSI data.
The spatial augmentations include randomly crop, resize, rotate, flip,
and blur. Since HSIs are usually captured with a low spatial resolution
while their discriminative information is mainly derived from spectra,
these spatial transformations would not destroy their inherent semantics.
In contrast, the spectral augmentations should not include
strong transformations. In the paper, we adopt random band permutation
with adjacent grouping and random band erasure. It should be noted
adjacent permutation follows the continuity of spectral bands and
random band erasure plays a band selection role. Furthermore, we conduct
data augmentation by the composition of multiple transformations, which
has proven to be helpful in increasing the robustness of contrastive learning
models.

\subsubsection{Network Architecture}

The backbone network $f_{\theta}\left(\cdot\right)$ used in this
paper consists of a modified ResNet-18 network \cite{ResNet-He-CVPR-16}
that removes pooling layers and the classification layer. Nonetheless,
one can use any off-the-shelf models to implement it. The projection
head $g_{\phi}\left(\cdot\right)$ used in our experiment is comprised
of two fully connected layers with $512$ and $C$ output units, respectively.
The first layer of the projection head is followed by rectified linear
units, while the last layer is followed by softmax units.

\subsubsection{Optimization}

Our SSCC model is optimized using the Adam optimizer with a batch size
of $512$ and a weight decay parameter of $5\cdot10^{-3}$. The learning
rate starts at $0.02$ and is linearly decreased with a scale of $0.1$
for every $20$ epochs. We run a search for the trade-off parameter
$\alpha$ and $\lambda$ of the loss function and found the best results
for $\alpha=5\cdot10^{-3}$ and $\lambda=5\cdot10^{-2}$.
We follow the SimCLR \cite{SSL-SimCLR-Chen-ICML-20} and set the temperature
parameters $\tau$ to $0.5$. For computational efficiency, we reduce
the initial spectral channels into $8$ using principal component
analysis. We use a local region of $13\times13$ pixels surrounding
each sample to generate training HSI cells. The training is conducted
on an NVIDIA TITAN XP GPU with 11 GB of graphic memory. 
\begin{table*}[htbp!]   
	\centering   
	\caption{Groundtruth classes for Indian Pines, Houston and Salinas datasets and their respective samples number.}     
	\begin{tabular}{llc|lc|lc}     
		\toprule     
		\multicolumn{1}{c}{\multirow{2}[4]{*}{Class No.}} & \multicolumn{2}{c}{Indian Pines} & \multicolumn{2}{c}{Houston} & \multicolumn{2}{c}{Salinas} \\ 
		\cmidrule(r){2-3}  \cmidrule(r){4-5}  \cmidrule(r){6-7}          
		& Class name & \multicolumn{1}{l}{\# samples} & \multicolumn{1}{l}{Class name} & \multicolumn{1}{l}{\# samples} & Class name & \multicolumn{1}{l}{\# samples} \\     
		\midrule     1     & Alfalfa & 46    & \multicolumn{1}{l}{Healthy grass} & 1251  & Brocoli\_green\_weeds\_1 & 2009 \\     2     & Corn-notill & 1428  & \multicolumn{1}{l}{Stressed grass} & 1254  & Brocoli\_green\_weeds\_2 & 3726 \\     3     & Corn-mintill & 830   & \multicolumn{1}{l}{Synthetic grass} & 697   & Fallow & 1976 \\     4     & Corn  & 237   & \multicolumn{1}{l}{Trees} & 1244  & Fallow\_rough\_plow & 1394 \\     5     & Grass-pasture & 483   & \multicolumn{1}{l}{ Soil} & 1242  & Fallow\_smooth & 2678 \\     6     & Grass-trees & 730   & \multicolumn{1}{l}{Water} & 325   & Stubble & 3959 \\     7     & Grass-pasture-mowed & 28    & \multicolumn{1}{l}{ Residential} & 1268  & Celery & 3579 \\     8     & Hay-windrowed & 478   & \multicolumn{1}{l}{ Commercial} & 1244  & Grapes\_untrained & 11271 \\     9     & Oats  & 20    & \multicolumn{1}{l}{ Road} & 1252  & Soil\_vinyard\_develop & 6203 \\     10    & Soybean-notill & 972   & \multicolumn{1}{l}{ Highway} & 1227  & Corn\_senesced\_green\_weeds & 3278 \\     11    & Soybean-mintill & 2455  & \multicolumn{1}{l}{Railway} & 1235  & Lettuce\_romaine\_4wk & 1068 \\     12    & Soybean-clean & 593   & \multicolumn{1}{l}{ Parking Lot 1} & 1233  & Lettuce\_romaine\_5wk & 1927 \\     13    & Wheat & 205   & \multicolumn{1}{l}{Parking Lot 2} & 469   & Lettuce\_romaine\_6wk & 916 \\     14    & Woods & 1265  & \multicolumn{1}{l}{Tennis Court} & 428   & Lettuce\_romaine\_7wk & 1070 \\     15    & Buildings-Grass-Trees-Drives & 386   & \multicolumn{1}{l}{Running Track} & 660   & Vinyard\_untrained & 7268 \\     16    & Stone-Steel-Towers & 93    &       &       & Vinyard\_vertical\_trellis & 1807 \\     \bottomrule     \end{tabular}  
	\label{tab:dataset} 
\end{table*}

\section{Results\label{sec:Results}}

\subsection{Experiment Setup}

\subsubsection{Datasets}

We conduct experiments on three real HSI benchmark datasets that are
commonly used in HSI classification: Indian Pines, Houston, and Salinas.
The Indian Pines and Salinas datasets\footnote{http://www.ehu.eus/ccwintco/index.php?title=\\Hyperspectral\_Remote\_Sensing\_Scenes}
were collected by the 224-band AVIRIS sensor over the Indian Pines
test site in North-western Indiana and Salinas Valley, California,
respectively. Both datasets include 16 land cover types and their
spatial sizes are $145\times145$ and $512\times217$, respectively.
By removing bands covering the region of water absorption and noise,
their spectral bands are 200 and 204, respectively. The Houston dataset
was originally used for the 2013 IEEE GRSS data fusion contest\footnote{http://www.grss-ieee.org/community/technical-committees/data- fusion/2013-ieee-grss-data-fusion-contest/},
and collected using the ITRES CASI-1500 sensor over the campus of
University of Houston and its surrounding rural areas in TX, USA.
The dataset has $144$ spectral bands in the 380 nm to 1050 nm region
and comprises $349\times1905$ pixels and 15 classes. More details
on the datasets are described in Table \ref{tab:dataset}. Following
\cite{HSIClu-Review-ZHai-GSM-21,HSIClu-GCSC-CAI-TGRS-20}, we assume
that the number of clusters in these HSI datasets is known.

\begin{table*}[htbp]   
\centering   
\caption{Comparison of clustering performance on Indian Pines. The best results are shown in bold.}
\resizebox{2.\columnwidth}{!}{     
\begin{tabular}{cccccccccccccc}     \toprule     
No. & $k$-means \cite{kmeans-TPAMI-02}& FCM \cite{FCM-ICFS-09}  & SC \cite{SpectralClu-NIPS-02}   & LSR \cite{Subspace-LSR-ECCV-12}   & ESC \cite{Subspace-ESC-ECCV18}   & HESSC \cite{HSIClu-HESSC-Kasra-RS-20} & S5C \cite{Subspace-S5C-NIPS-19}   & JSCC \cite{HSIClu-JSSC-ZHai-TGRS-21} & GCSSC \cite{HSIClu-GCSSC-Huang-TGRS-21}  & AE\cite{AE-Hinton-Sci-06}+$k$-means & DEC \cite{DeepClu-DEC-Xie-ICML-16}  & CC \cite{SSL-CC-Li-AAAI-21}   & SSCC \\     
\midrule     1     & 0.0000 & 0.0000 & 0.0000 & 0.0217 & 0.2826 & 0.0000 & 0.0000 & 0.1348 & 0.0000 & 0.0870 & 0.0000 & 0.0870 & 0.0000 \\     2     & 0.2108 & 0.0728 & 0.0273 & 0.1176 & 0.3165 & 0.2787 & 0.1821 & 0.3340 & 0.6107 & 0.1408 & 0.1877 & 0.2955 & 0.6422 \\     3     & 0.4904 & 0.0590 & 0.0048 & 0.1120 & 0.0446 & 0.2627 & 0.2699 & 0.3017 & 0.4031 & 0.1940 & 0.2892 & 0.5795 & 0.5783 \\     4     & 0.1603 & 0.1688 & 0.0000 & 0.2194 & 0.0253 & 0.3376 & 0.2236 & 0.1612 & 0.1111 & 0.1730 & 0.2405 & 0.8861 & 1.0000 \\     5     & 0.0000 & 0.0000 & 0.0166 & 0.2257 & 0.2526 & 0.5466 & 0.6439 & 0.5793 & 0.7841 & 0.5217 & 0.0000 & 0.6584 & 0.6584 \\     6     & 0.9411 & 0.2493 & 0.0000 & 0.8493 & 0.9014 & 0.3041 & 0.8986 & 0.5597 & 0.9742 & 0.9329 & 0.9726 & 0.9795 & 0.9945 \\     7     & 0.0000 & 0.0000 & 0.0000 & 0.0714 & 0.0714 & 0.9286 & 0.7143 & 0.0071 & 0.0000 & 0.0000 & 0.0000 & 0.0000 & 0.0000 \\     8     & 0.5879 & 1.0000 & 1.0000 & 0.5397 & 0.5523 & 0.9707 & 0.6611 & 0.7251 & 0.7496 & 0.7134 & 0.6464 & 1.0000 & 1.0000 \\     9     & 0.0000 & 0.0500 & 0.0000 & 0.0000 & 0.0000 & 0.0000 & 0.0500 & 0.1500 & 0.4340 & 0.0000 & 0.0000 & 0.0000 & 0.0000 \\     10    & 0.2027 & 0.0134 & 0.0031 & 0.1296 & 0.2387 & 0.2623 & 0.1564 & 0.4866 & 0.3853 & 0.5021 & 0.3693 & 0.4095 & 0.7572 \\     11    & 0.3593 & 0.9010 & 0.9609 & 0.1267 & 0.7193 & 0.1898 & 0.3283 & 0.5871 & 0.6908 & 0.3752 & 0.3202 & 0.2957 & 0.3572 \\     12    & 0.1484 & 0.0590 & 0.0017 & 0.1012 & 0.1265 & 0.1551 & 0.2361 & 0.1926 & 0.2798 & 0.1551 & 0.2749 & 0.7437 & 0.4182 \\     13    & 0.0000 & 0.8878 & 0.1463 & 0.8098 & 0.7463 & 0.4878 & 0.7317 & 0.6127 & 1.0000 & 0.5024 & 0.3610 & 1.0000 & 1.0000 \\     14    & 0.7233 & 0.7534 & 1.0000 & 0.7731 & 0.7486 & 0.3621 & 0.6806 & 0.7031 & 0.6959 & 0.7605 & 0.7763 & 0.6854 & 0.7146 \\     15    & 0.7642 & 0.0000 & 0.0000 & 0.0674 & 0.3549 & 0.1969 & 0.0130 & 0.2549 & 0.7108 & 0.6995 & 0.7021 & 1.0000 & 0.8705 \\     16    & 0.6022 & 0.0000 & 0.7419 & 0.0968 & 1.0000 & 0.0215 & 1.0000 & 0.3763 & 0.0000 & 0.5591 & 0.6022 & 0.0000 & 0.0000 \\     \midrule     ACC   & 0.4046 & 0.4146 & 0.4153 & 0.2907 & 0.4837 & 0.3045 & 0.3950 & 0.4866 & 0.5309 & 0.4458 & 0.4171 & 0.5514 & \textbf{0.6305} \\     Kappa & 0.3405 & 0.3016 & 0.2794 & 0.2384 & 0.4018 & 0.2520 & 0.3396 & 0.4254 & N/A    & 0.3847 & 0.3591 & 0.5192 & \textbf{0.5985} \\     NMI   & 0.4690 & 0.4102 & 0.5041 & 0.3065 & 0.5022 & 0.3983 & 0.4437 & N/A    & 0.5638 & 0.4527 & 0.4787 & 0.6644 & \textbf{0.7136} \\     ARI   & 0.2670 & 0.2492 & 0.2455 & 0.1747 & 0.3118 & 0.1746 & 0.2555 & N/A    & 0.3651 & 0.2661 & 0.2968 & 0.4070 & \textbf{0.4867} \\     Purity & 0.5437 & 0.4336 & 0.4226 & 0.4523 & 0.5464 & 0.4978 & 0.5278 & 0.5689 & N/A    & 0.5516 & 0.5470 & 0.7429 & \textbf{0.8079} \\     \bottomrule     
\end{tabular}
}  
\label{tab:acc-inp}
\end{table*}

\begin{table*}[htbp]   
\centering   
\caption{Comparison of clustering performance on Houston. The best results are shown in bold.}
\resizebox{2.\columnwidth}{!}{      
\begin{tabular}{cccccccccccccc}     
\toprule     
No. & $k$-means \cite{kmeans-TPAMI-02}& FCM \cite{FCM-ICFS-09}  & SC \cite{SpectralClu-NIPS-02}   & LSR \cite{Subspace-LSR-ECCV-12}   & ESC \cite{Subspace-ESC-ECCV18}   & HESSC \cite{HSIClu-HESSC-Kasra-RS-20} & S5C \cite{Subspace-S5C-NIPS-19}   & JSCC \cite{HSIClu-JSSC-ZHai-TGRS-21} & GCSSC \cite{HSIClu-GCSSC-Huang-TGRS-21}  & AE\cite{AE-Hinton-Sci-06}+$k$-means & DEC \cite{DeepClu-DEC-Xie-ICML-16}  & CC \cite{SSL-CC-Li-AAAI-21}   & SSCC \\     \midrule     1     & 0.8177 & 0.4932 & 0.7786 & 0.8545 & 0.2894 & 0.4988 & 0.8137 & N/A    & N/A    & 0.8129 & 0.8185 & 0.7506 & 0.8233 \\     2     & 0.1085 & 0.2831 & 0.4928 & 0.2448 & 0.3692 & 0.3365 & 0.5287 & N/A    & N/A    & 0.0973 & 0.7783 & 0.6938 & 0.5909 \\     3     & 0.7575 & 0.0000 & 0.8651 & 0.8522 & 0.9326 & 0.9742 & 0.8881 & N/A    & N/A    & 0.9842 & 0.4390 & 0.9900 & 1.0000 \\     4     & 0.7130 & 0.3834 & 0.2846 & 0.0217 & 0.9437 & 0.6214 & 0.6543 & N/A    & N/A    & 0.7436 & 0.0000 & 0.6624 & 0.7870 \\     5     & 0.7931 & 0.6932 & 0.7440 & 0.7568 & 0.8019 & 0.9879 & 0.8623 & N/A    & N/A    & 0.8285 & 0.9815 & 0.7198 & 0.7238 \\     6     & 0.0000 & 0.0185 & 0.0831 & 0.1662 & 0.7815 & 0.0000 & 0.0554 & N/A    & N/A    & 0.7600 & 0.0000 & 0.0000 & 0.8677 \\     7     & 0.0000 & 0.0789 & 0.3793 & 0.1656 & 0.9196 & 0.0804 & 0.3604 & N/A    & N/A    & 0.8028 & 0.0000 & 0.6088 & 0.5994 \\     8     & 0.2251 & 0.0949 & 0.4526 & 0.4043 & 0.1969 & 0.3199 & 0.3031 & N/A    & N/A    & 0.2291 & 0.2098 & 0.3778 & 0.1447 \\     9     & 0.0000 & 0.1070 & 0.3986 & 0.1366 & 0.7939 & 0.7819 & 0.3514 & N/A    & N/A    & 0.0000 & 0.4744 & 0.3011 & 0.4281 \\     10    & 0.2958 & 0.2347 & 0.3195 & 0.0000 & 0.3121 & 0.0000 & 0.2738 & N/A    & N/A    & 0.2771 & 0.6544 & 0.3325 & 0.3822 \\     11    & 0.2785 & 0.5417 & 0.4332 & 0.3700 & 0.0008 & 0.3279 & 0.2680 & N/A    & N/A    & 0.2777 & 0.0089 & 0.2874 & 0.4470 \\     12    & 0.6975 & 0.8021 & 0.3552 & 0.4436 & 0.2141 & 0.1249 & 0.3131 & N/A    & N/A    & 0.7875 & 0.5929 & 0.4039 & 0.2717 \\     13    & 0.0000 & 0.0000 & 0.0512 & 0.0000 & 0.0149 & 0.0085 & 0.9275 & N/A    & N/A    & 0.0000 & 0.6482 & 0.6077 & 0.8188 \\     14    & 0.9977 & 0.0000 & 0.0000 & 0.3014 & 0.0000 & 0.0000 & 0.9393 & N/A    & N/A    & 0.9907 & 1.0000 & 0.0000 & 0.0000 \\     15    & 0.3742 & 0.0000 & 1.0000 & 0.6379 & 1.0000 & 0.0000 & 0.6061 & N/A    & N/A    & 0.5727 & 0.6258 & 0.9924 & 1.0000 \\     \midrule     ACC   & 0.4046 & 0.3070 & 0.4719 & 0.3612 & 0.5070 & 0.3837 & 0.5168 & N/A    & N/A    & 0.5180 & 0.4704 & 0.5348 & \textbf{0.5658} \\     Kappa & 0.3557 & 0.2456 & 0.4336 & 0.3113 & 0.4691 & 0.3354 & 0.4809 & N/A    & N/A    & 0.4783 & 0.4278 & 0.5008 & \textbf{0.5341} \\     NMI   & 0.5317 & 0.3623 & 0.5510 & 0.4529 & 0.6167 & 0.4582 & 0.5458 & N/A    & N/A    & 0.5871 & 0.5660 & 0.5949 & \textbf{0.6349} \\     ARI   & 0.2592 & 0.1488 & 0.3390 & 0.2515 & 0.3584 & 0.2282 & 0.3468 & N/A    & N/A    & 0.3613 & 0.3330 & 0.3977 & \textbf{0.4266} \\     Purity & 0.4595 & 0.3272 & 0.5356 & 0.3847 & 0.5730 & 0.4326 & 0.5356 & N/A    & N/A    & 0.5460 & 0.5081 & 0.5943 & \textbf{0.6043} \\     
\bottomrule     
\end{tabular}
}
\label{tab:acc-hou}
\end{table*}

\begin{table*}[htbp]   
\centering   
\caption{Comparison of clustering performance on Salinas. The best results are shown in bold. It should be noted that both SC and LSR suffer from out-of-memory (OOM) on this dataset caused by the construction of a $N\times N$ affinity matrix.}  
\resizebox{2.\columnwidth}{!}{   
\begin{tabular}{cccccccccccccc}     
\toprule     
No. & $k$-means \cite{kmeans-TPAMI-02}& FCM \cite{FCM-ICFS-09}  & SC \cite{SpectralClu-NIPS-02}   & LSR \cite{Subspace-LSR-ECCV-12}   & ESC \cite{Subspace-ESC-ECCV18}   & HESSC \cite{HSIClu-HESSC-Kasra-RS-20} & S5C \cite{Subspace-S5C-NIPS-19}   & JSCC \cite{HSIClu-JSSC-ZHai-TGRS-21} & GCSSC \cite{HSIClu-GCSSC-Huang-TGRS-21}  & AE\cite{AE-Hinton-Sci-06}+$k$-means & DEC \cite{DeepClu-DEC-Xie-ICML-16}  & CC \cite{SSL-CC-Li-AAAI-21}   & SSCC \\     \midrule     1     & 0.0000 & 0.9278 & OOM    & OOM    & 0.0000 & 0.9278 & 0.0000 & 0.3996 & 0.7878 & 0.0000 & 0.9000 & 1.0000 & 0.9985 \\     2     & 0.9979 & 0.1119 & OOM    & OOM    & 0.9938 & 0.6887 & 0.9844 & 0.9443 & 1.0000 & 1.0000 & 0.7933 & 0.8969 & 0.9775 \\     3     & 0.5951 & 0.0000 & OOM    & OOM    & 0.2915 & 0.0000 & 0.1204 & 0.0764 & 0.0000 & 0.5501 & 0.0000 & 1.0000 & 0.9408 \\     4     & 1.0000 & 0.0000 & OOM    & OOM    & 0.9964 & 0.0000 & 0.9663 & 0.9980 & 0.8969 & 0.9340 & 0.5079 & 0.0000 & 0.0000 \\     5     & 0.2521 & 0.2670 & OOM    & OOM    & 0.8857 & 0.9899 & 0.7976 & 0.6843 & 0.7052 & 0.7367 & 0.7412 & 0.5310 & 0.9642 \\     6     & 0.8032 & 0.0000 & OOM    & OOM    & 0.9194 & 0.5792 & 0.8894 & 0.9695 & 0.9875 & 0.8406 & 0.8366 & 0.6603 & 1.0000 \\     7     & 0.7058 & 0.8055 & OOM    & OOM    & 0.9486 & 0.7726 & 0.8838 & 0.9922 & 0.9992 & 0.3328 & 0.6943 & 0.7499 & 0.9994 \\     8     & 0.7206 & 0.7828 & OOM    & OOM    & 0.9378 & 0.5955 & 0.2700 & 0.7724 & 0.6477 & 0.5008 & 0.8565 & 0.3615 & 0.4416 \\     9     & 1.0000 & 0.9919 & OOM    & OOM    & 0.9886 & 0.9321 & 0.3082 & 0.7789 & 0.9955 & 0.9997 & 0.9994 & 0.7201 & 1.0000 \\     10    & 0.6135 & 0.4549 & OOM    & OOM    & 0.2123 & 0.1312 & 0.5018 & 0.1160 & 0.7166 & 0.5863 & 0.3350 & 0.9451 & 0.9170 \\     11    & 0.0000 & 0.0000 & OOM    & OOM    & 0.0000 & 0.0000 & 0.3970 & 0.7562 & 0.8471 & 0.0000 & 0.0000 & 1.0000 & 0.0000 \\     12    & 0.9958 & 0.5864 & OOM    & OOM    & 0.8895 & 0.0000 & 0.3752 & 0.5103 & 0.8772 & 0.9460 & 0.8293 & 0.9984 & 1.0000 \\     13    & 0.0000 & 0.0000 & OOM    & OOM    & 0.0011 & 0.9214 & 0.0000 & 0.9314 & 0.0000 & 0.0000 & 0.0000 & 0.0000 & 0.0000 \\     14    & 1.0000 & 0.0009 & OOM    & OOM    & 0.9607 & 0.8131 & 0.2794 & 0.8234 & 0.5062 & 0.9645 & 0.0000 & 0.0000 & 1.0000 \\     15    & 0.1977 & 0.2369 & OOM    & OOM    & 0.0003 & 0.1560 & 0.2720 & 0.5191 & 0.5017 & 0.6237 & 0.0579 & 0.5687 & 0.8687 \\     16    & 0.9524 & 0.7227 & OOM    & OOM    & 0.8866 & 0.4079 & 0.8379 & 0.5717 & 0.9954 & 0.9917 & 0.9092 & 1.0000 & 0.7222 \\     \midrule     ACC   & 0.6494 & 0.4897 & OOM    & OOM    & 0.6802 & 0.5293 & 0.4731 & 0.6897 & 0.7666 & 0.6568 & 0.6257 & 0.6396 & \textbf{0.7838} \\     Kappa & 0.6087 & 0.4206 & OOM    & OOM    & 0.6385 & 0.4785 & 0.4318 & 0.6548 & N/A    & 0.6206 & 0.5796 & 0.6123 & \textbf{0.7611} \\     NMI   & 0.8109 & 0.6085 & OOM    & OOM    & 0.7938 & 0.6238 & 0.6186 & N/A    & 0.8311 & 0.7720 & 0.7705 & 0.7806 & \textbf{0.8705} \\     ARI   & 0.5981 & 0.4006 & OOM    & OOM    & 0.6185 & 0.4150 & 0.3738 & N/A    & 0.6722 & 0.5548 & 0.5824 & 0.5330 & \textbf{0.6682} \\     Purity & 0.7383 & 0.4957 & OOM    & OOM    & 0.6981 & 0.5881 & 0.6324 & 0.7115 & N/A    & 0.7272 & 0.6917 & 0.7943 & \textbf{0.7946} \\     
\bottomrule     
\end{tabular}
}
\label{tab:acc-san}
\end{table*}

\subsubsection{Evaluation Metrics}

Five popular metrics \cite{HSIClu-Review-ZHai-GSM-21,SSL-CC-Li-AAAI-21,Subspace-S5C-NIPS-19,HSIClu-GR-RSCNet-CAI-INS-21}
are used to quantitatively evaluate the clustering performance across
different perspectives: overall accuracy (ACC), Kappa coefficient
(Kappa), normalized mutual information (NMI), adjusted rand index
(ARI), and Purity. These metrics range in $\left[0,1\right]$ (ACC,
NMI, and Purity) or $\left[-1,1\right]$ (Kappa and ARI), and higher
scores signify more accurate clustering results are achieved. Following
the common process of clustering evaluation, we adopt the Hungarian
algorithm \cite{Hungarian-1955} to match the predicted labels to
the ground truth before calculating metrics.

\begin{table*}[htbp!]    	
	\centering    	
	\caption{Ablation study on the components of the loss function and different parameters, where $\mathcal{L}_{\mathcal{W}}$ and $\mathcal{L}_{\mathcal{B}}$ indicate the InfoNCE loss and the Barlow Twins loss, respectively, and '\cmark' means the corresponding  loss is used while '\xmark' is reverse. We also report the results obtained under different hyperparameter settings of $\tau$ or $\lambda$.} 	
	\resizebox{2.\columnwidth}{!}{      	
		\begin{tabular}{cccccccccccccccccc}      		
			\toprule      		
			\multicolumn{2}{c}{Components} & Settings & \multicolumn{5}{c}{Indian Pines} & \multicolumn{5}{c}{Houston} & \multicolumn{5}{c}{Salinas} \\      		
			\cmidrule(r){1-2} \cmidrule(r){3-3} \cmidrule(r){4-8}  \cmidrule(r){9-13} \cmidrule(r){14-18}   
			$\mathcal{L}_{\mathcal{W}}$ & $\mathcal{L}_{\mathcal{B}}$ & $\tau$ / $\lambda$ & ACC & Kappa & NMI & ARI & Purity & ACC & Kappa & NMI & ARI & Purity & ACC & Kappa & NMI & ARI & Purity \\      		
			\midrule     		 
			\cmark & \xmark   & $ \tau=0.05 $ & 0.2482 & 0.1964 & 0.3116 & 0.1113 & 0.3913 & 0.2508 & 0.1960 & 0.2393 & 0.0907 & 0.2629 & 0.2177 & 0.1652 & 0.2777 & 0.0993 & 0.3522 \\     
			\cmark & \xmark   & $ \tau=0.10 $ & 0.2576 & 0.2102 & 0.3287 & 0.1176 & 0.4112 & 0.2577 & 0.2052 & 0.2867 & 0.1082 & 0.2741 & 0.2701 & 0.2213 & 0.3489 & 0.1247 & 0.3557 \\     
			\cmark & \xmark   & $ \tau=0.50 $ & 0.5678 & 0.5337 & 0.6781 & 0.4304 & 0.7347 & 0.5348 & 0.5002 & 0.6274 & 0.4226 & 0.5729 & 0.6208 & 0.5885 & 0.8214 & 0.6062 & 0.7531 \\     
			\cmark & \xmark   & $ \tau=1.00 $   & 0.5712 & 0.5176 & 0.6036 & 0.3884 & 0.6215 & 0.0202 & -0.0525 & 0.5614 & 0.3615 & 0.4702 & 0.5213 & 0.4629 & 0.7093 & 0.4556 & 0.5253 \\      		 
			\midrule      		 
			\xmark   & \cmark & $ \lambda=0.005 $ & 0.5243 & 0.4837 & 0.5721 & 0.3556 & 0.6656 & 0.3715 & 0.3215 & 0.4561 & 0.2166 & 0.4381 & 0.6883 & 0.6562 & 0.8217 & 0.5976 & 0.7463 \\     
			\xmark   & \cmark & $ \lambda=0.050 $ & 0.5619 & 0.5281 & 0.6637 & 0.4279 & 0.7334 & 0.4152 & 0.3719 & 0.5073 & 0.2932 & 0.4926 & 0.7367 & 0.7109 & 0.8615 & 0.6372 & 0.7956 \\     
			\xmark   & \cmark & $ \lambda=1.000 $ & 0.4633 & 0.4064 & 0.5591 & 0.2668 & 0.6378 & 0.5230 & 0.4861 & 0.5559 & 0.3511 & 0.5668 & 0.6920 & 0.6614 & 0.8148 & 0.5656 & 0.7636 \\    
			\xmark   & \cmark & $ \lambda=50.00 $ & 0.0468 & 0.0077 & 0.0891 & 0.0252 & 0.2513 & 0.0217 & -0.0006 & 0.0179 & 0.0004 & 0.0923 & 0.0366 & -0.0186 & 0.3657 & 0.1893 & 0.3228 \\      		 
			\midrule      
			\cmark & $\mathcal{L}_{\mathcal{W}}$ & 0.50 / 0.50 & 0.5601 & 0.5207 & 0.6576 & 0.4064 & 0.6701 & 0.4342 & 0.3937 & 0.5663 & 0.3174 & 0.5113 & 0     & -0.0613 & 0.8183 & 0.5908 & 0.7244 \\     
			$\mathcal{L}_{\mathcal{B}}$ & \cmark & 0.05 / 0.05 & 0.5417 & 0.5066 & 0.6441 & 0.4091 & 0.7180 & 0.3652 & 0.3183 & 0.5212 & 0.2662 & 0.4825 & 0.6974 & 0.6701 & 0.8330 & 0.6272 & 0.8272 \\      
			\midrule     		  
			\cmark & \cmark & 0.50 / 0.05 & \textbf{0.6270} & \textbf{0.5950} & \textbf{0.7117} & \textbf{0.4907} & \textbf{0.8090} & \textbf{0.5658} & \textbf{0.5341} & \textbf{0.6349} & \textbf{0.4266} & \textbf{0.6043} & \textbf{0.7487} & \textbf{0.7175} & \textbf{0.8566} & \textbf{0.6745} & \textbf{0.7598} \\      		 
			\bottomrule      	 
	\end{tabular}}   
	\label{tab:ablation-loss}  \end{table*}

\begin{table*}[htbp]     
	\centering    
	\caption{Clustering perfomance achieved with different batch sizes.}   
	\resizebox{2.\columnwidth}{!}{    
		\begin{tabular}{cccccccccccccccc}     	 
			\toprule      	 
			\multirow{2}[4]{*}{Batch size} & \multicolumn{5}{c}{Indian Pines}      & \multicolumn{5}{c}{Houston}          & \multicolumn{5}{c}{Salinas} \\  	 \cmidrule(r){2-6}  \cmidrule(r){7-11} \cmidrule(r){12-16}       & ACC   & Kappa & NMI   & ARI   & Purity & ACC   & Kappa & NMI   & ARI   & Purity & ACC   & Kappa & NMI   & ARI   & Purity \\      	 \midrule      64    & 0.5392 & 0.5044 & 0.6537 & 0.4217 & 0.7389 & 0.5165 & 0.4812 & 0.5850 & 0.3691 & 0.5742 & 0.6798 & 0.6550 & 0.8214 & 0.5828 & 0.8324 \\      	128   & 0.5608 & 0.5274 & 0.6684 & 0.4328 & 0.7471 & 0.5308 & 0.4976 & 0.6332 & 0.4119 & 0.5943 & 0.7170 & 0.6874 & 0.8449 & 0.6098 & 0.7753 \\      	256   & 0.5786 & 0.5470 & 0.6679 & 0.4388 & \textbf{0.7561} & 0.5339 & 0.5002 & \textbf{0.6424} & \textbf{0.4293} & \textbf{0.6089} & 0.7250 & 0.7033 & 0.8549 & 0.6237 & 0.8645 \\      	512   & \textbf{0.5942} & \textbf{0.5598} & \textbf{0.6782} & \textbf{0.4542} & 0.7462 & \textbf{0.5437} & \textbf{0.5104} & 0.6325 & 0.4153 & 0.6070 & \textbf{0.7435} & \textbf{0.7200 } & \textbf{0.8669 }& \textbf{0.6637} & \textbf{0.8657} \\      	 
			\bottomrule       
	\end{tabular} }   
	\label{tab:ablation-batchsize}  
\end{table*}

\begin{figure*}[htbp]
	\begin{minipage}[t]{1\columnwidth}
		\centering
		\includegraphics[width=1.1\columnwidth]{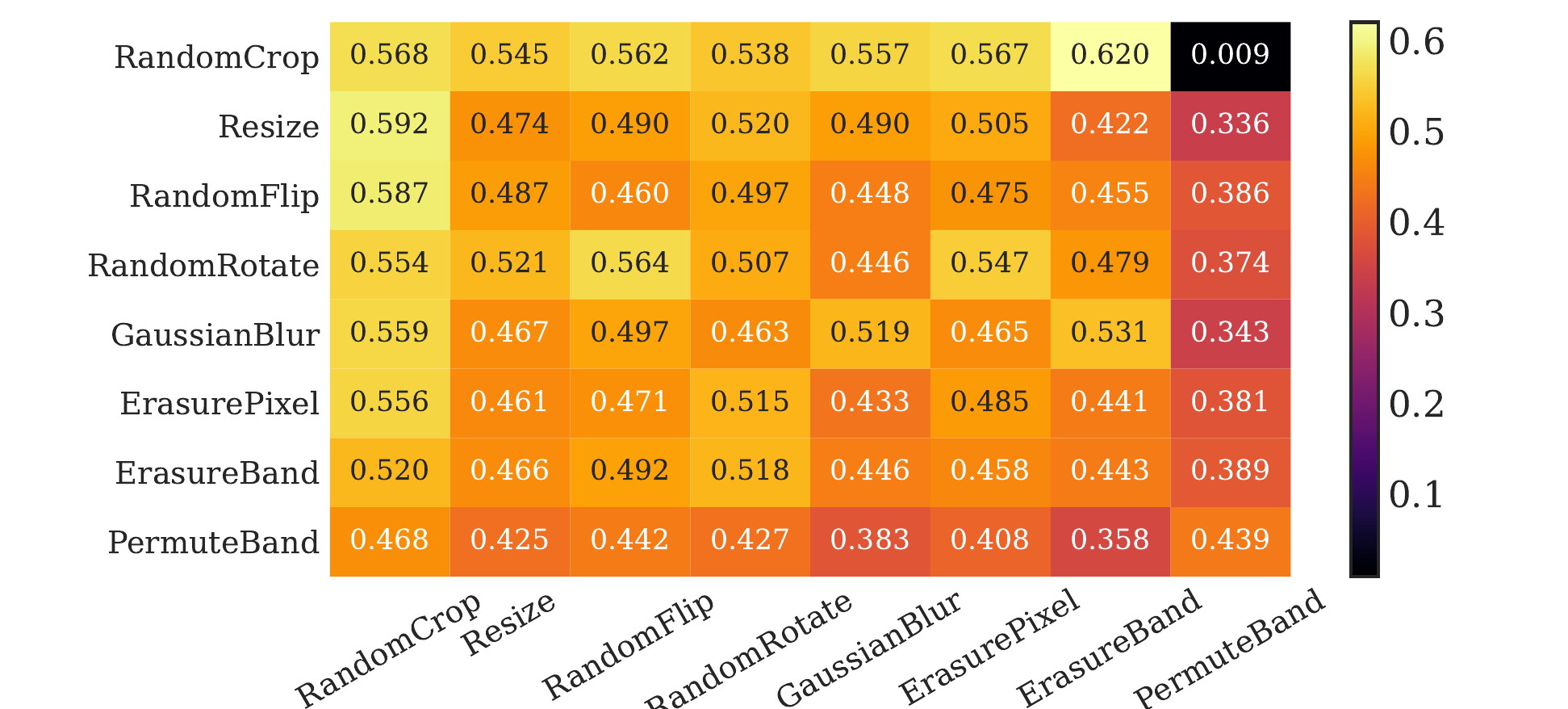}
		\caption{Evaluation under individual or composition of data augmentations on
			Indian Pines dataset. For all columns, diagonal entries correspond
			to a single transformation, and off-diagonals correspond to the composition
			of two transformations.\label{fig:aug}}
	\end{minipage}
	\begin{minipage}[t]{1\columnwidth}
		\centering
		\includegraphics[width=0.9\columnwidth]{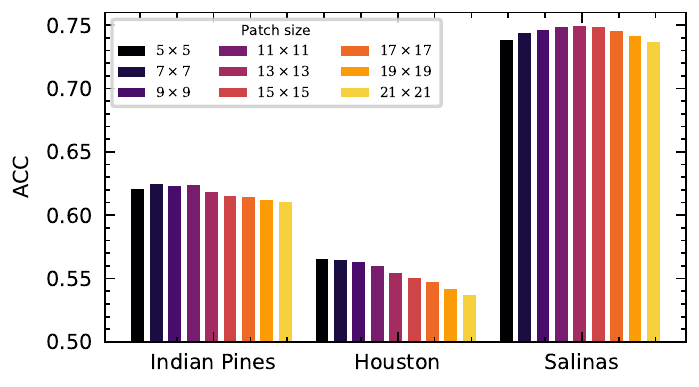}
		\caption{Clustering performance of SSCC with different input patch sizes. \label{fig:patch}}
	\end{minipage}
\end{figure*}

\subsection{Comparison with State-of-The-Art Methods}

We compare our SSCC model with $12$ representative state-of-the-art
clustering approaches, including $k$-means \cite{kmeans-TPAMI-02},
FCM \cite{FCM-ICFS-09}, spectral clustering (SC) \cite{SpectralClu-NIPS-02},
subspace clustering using least squares regression (LSR) \cite{Subspace-LSR-ECCV-12},
exemplar-based subspace clustering (ESC) \cite{Subspace-ESC-ECCV18},
hierarchical sparse subspace clustering (HESSC) \cite{HSIClu-HESSC-Kasra-RS-20},
selective sampling-based scalable sparse subspace clustering (S5C)
\cite{Subspace-S5C-NIPS-19}, joint sparse subspace clustering (JSCC)
\cite{HSIClu-JSSC-ZHai-TGRS-21}, graph convolutional sparse subspace
coclustering (GCSSC) \cite{HSIClu-GCSSC-Huang-TGRS-21}, AE \cite{AE-Hinton-Sci-06}+$k$-means,
DEC \cite{DeepClu-DEC-Xie-ICML-16}, and CC \cite{SSL-CC-Li-AAAI-21}.
Notably, we include six popular subspace clustering models, four of
which (i.e., ESC, HESSC, S5C, and JSCC) are scalable for large datasets.
Furthermore, AE+$k$-means, DEC, and CC belong to deep clustering
models. Except for JSCC \cite{HSIClu-JSSC-ZHai-TGRS-21} and GCSSC
\cite{HSIClu-GCSSC-Huang-TGRS-21}, all the compared methods are reproduced
under the same preprocessing and follow the settings suggested in
the corresponding official releases. Since there are no available codes
for JSCC and GCSSC, we compare their results reported in \cite{HSIClu-JSSC-ZHai-TGRS-21} and \cite{HSIClu-GCSSC-Huang-TGRS-21},
respectively.

In Table \ref{tab:acc-inp}, \ref{tab:acc-hou}, and \ref{tab:acc-san},
we report the comparison results on Indian Pines, Houston, and Salinas
datasets, respectively. We can obtain the following three conclusions.
First, our SSCC model consistently outperforms other compared methods
by large margins across all datasets. For example, ACCs obtained by
SSCC on three datasets are $63.05\%$, $56.58\%$, and $78.38\%$ respectively,
which improves upon the second-best model (CC on the Indian Pines
and Houston, and GCSSC on the Salinas) by a margin of $7.91\%$, $3.10\%$,
and $1.72\%$ (absolute differences). For the single cluster's accuracy
(a.k.a., users' accuracy), SSCC perfectly identifies $3/16$, $2/15$,
and $4/16$ land cover types separately on the three datasets. These
signify the effectiveness and superiority of our model. 

Second, deep clustering models (e.g., AE+$k$-means, DEC, CC, and
SSCC) remarkably outperform shallow clustering models (e.g., $k$-means,
FCM, and SC). A conceivable reason is that shallow models measure
interrelation between raw data directly, thus failing to capture high-level
semantic and leading to poor robustness and accuracy. Nonetheless,
these classic clustering models can be significantly improved by combining
them with deep feature learning. For example, compared to the classic
$k$-means, the naive version of AE+$k$-means and its end-to-end
version (i.e., DEC) achieve a considerable improvement in terms of
clustering accuracy.

Third, SSCC shows better robustness than CC and better scalability
than offline clustering models. Compared to CC, the proposed SSCC
contains only one prediction head and its architecture is entirely
symmetric. Nonetheless, SSCC improves over CC with impressive margins
in terms of clustering accuracy across three datasets. This shows
that a single projection head and symmetric architecture can also
achieve better results by optimizing the proposed loss functions. Compared to offline
models (e.g., subspace clustering models), SSCC has an obvious advantage,
i.e., 'train once, run everywhere.' It means SSCC would not be restricted
by data sizes and training data. Besides, we can see that out-of-memory
(OOM) occurs in SC and LSR in Table \ref{tab:acc-san}. This is because
both of these methods require computation of a $N\times N$ affinity
matrix, which results in quadratic time and space complexity. 

\begin{figure*}[tbh]
	\resizebox{2.\columnwidth}{!}{
		\begin{centering}
			\subfloat[]{\includegraphics[width=0.5\columnwidth]{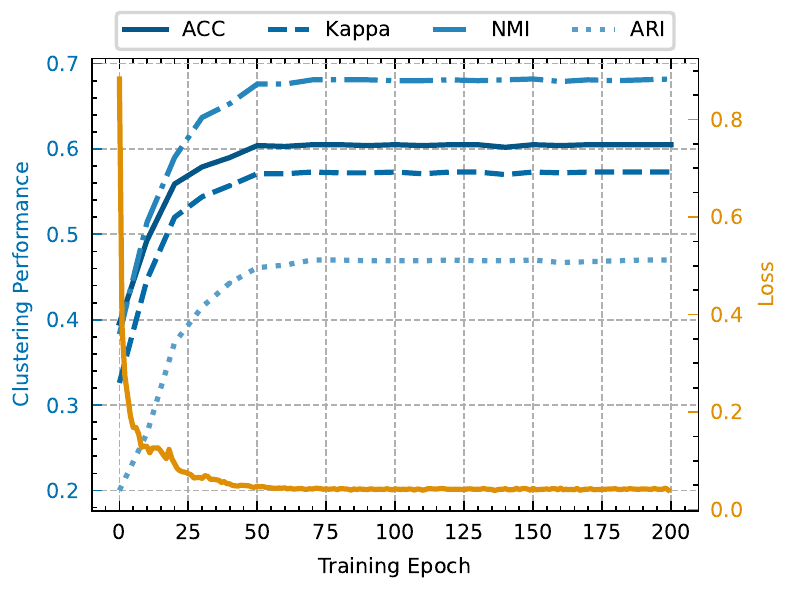}
				
			}\subfloat[]{\includegraphics[width=0.5\columnwidth]{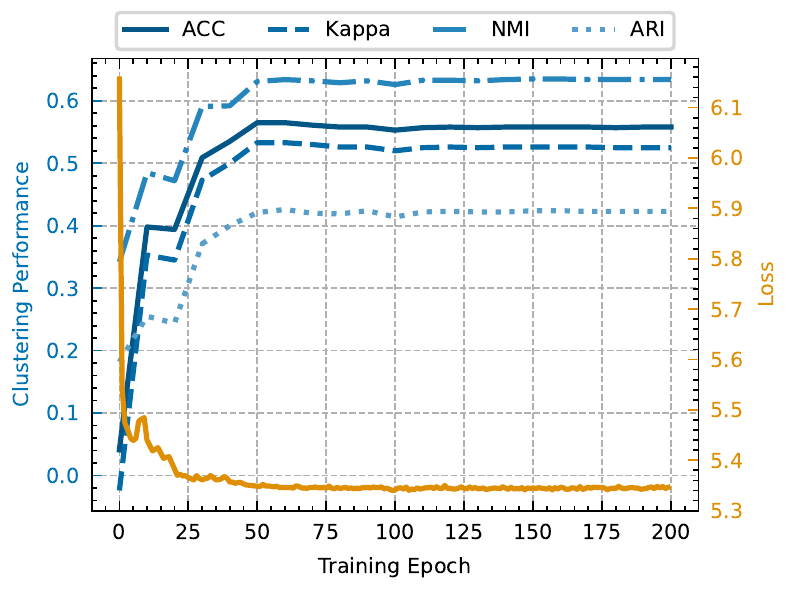}
				
			}\subfloat[]{\includegraphics[width=0.5\columnwidth]{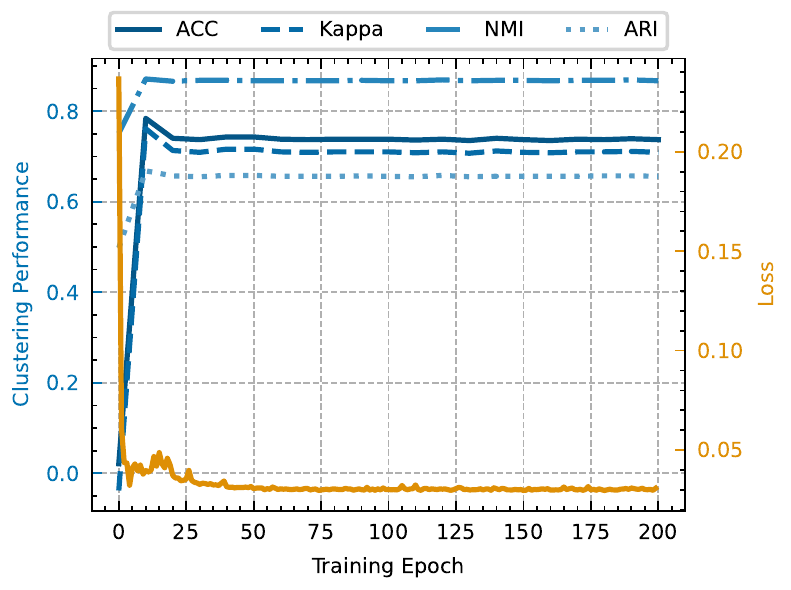}
				
			}\subfloat[]{\includegraphics[width=0.5\columnwidth]{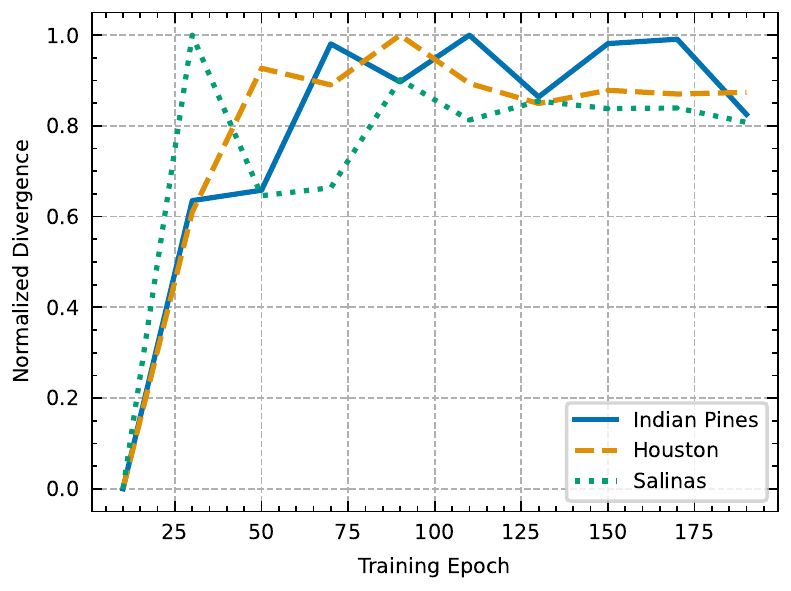}
				
			}
			\par\end{centering}
	}
	\caption{(a)-(c) Clustering performance of SSCC with increasing epochs on Indian
		Pines, Houston and Salinas datasets, respectively, where the x-axis
		denotes the training epoch, the left and right y-axis denote the clustering
		performance and the corresponding loss value. (d) Quantitative evaluation
		on the model discriminant power with respect to different epochs by computing
		the divergence of label representation.\label{fig:loss-acc}}
\end{figure*}

\begin{figure*}[tbh]
	\resizebox{2.\columnwidth}{!}{
		\includegraphics[width=0.4\columnwidth]{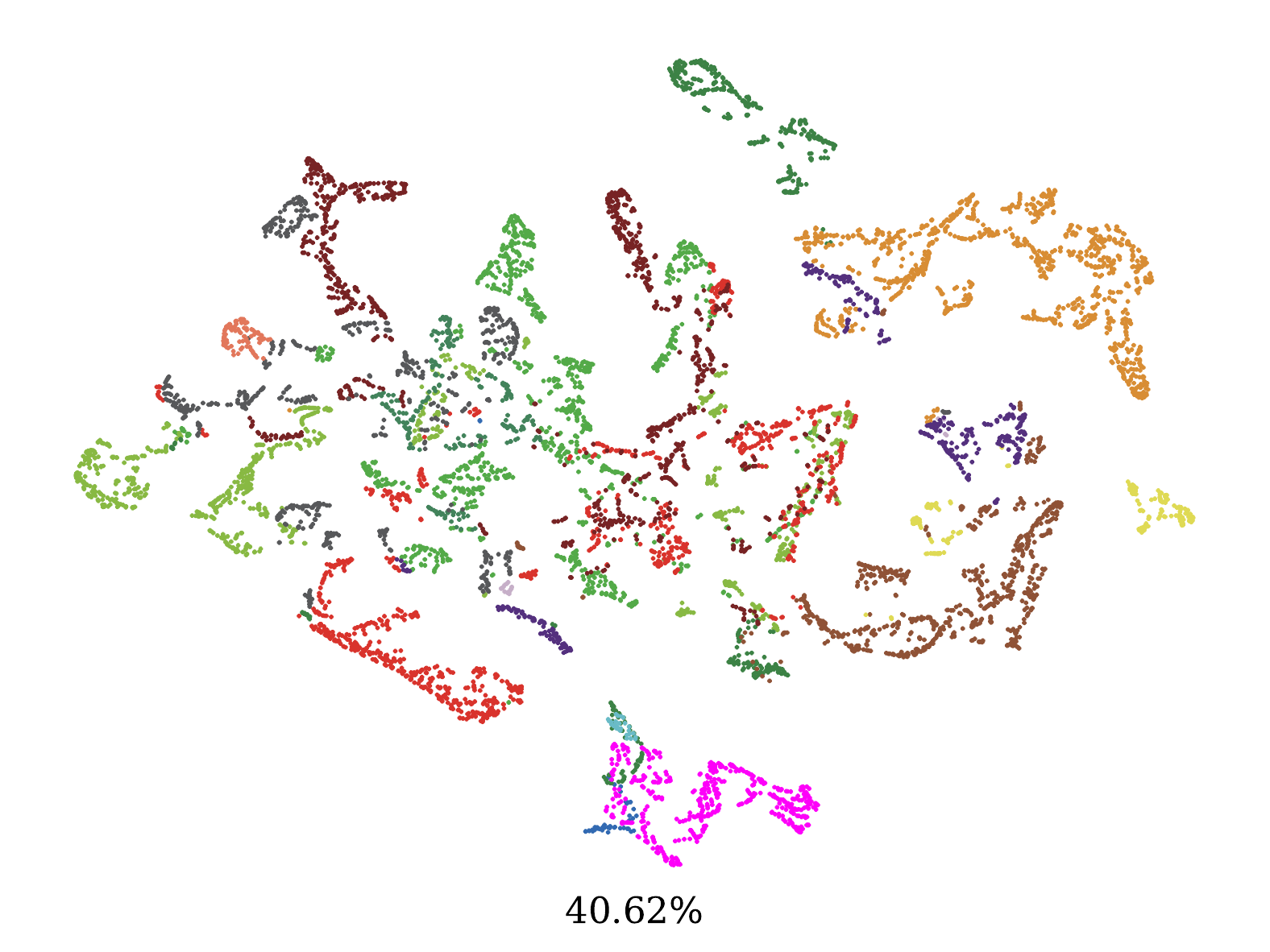}\includegraphics[width=0.4\columnwidth]{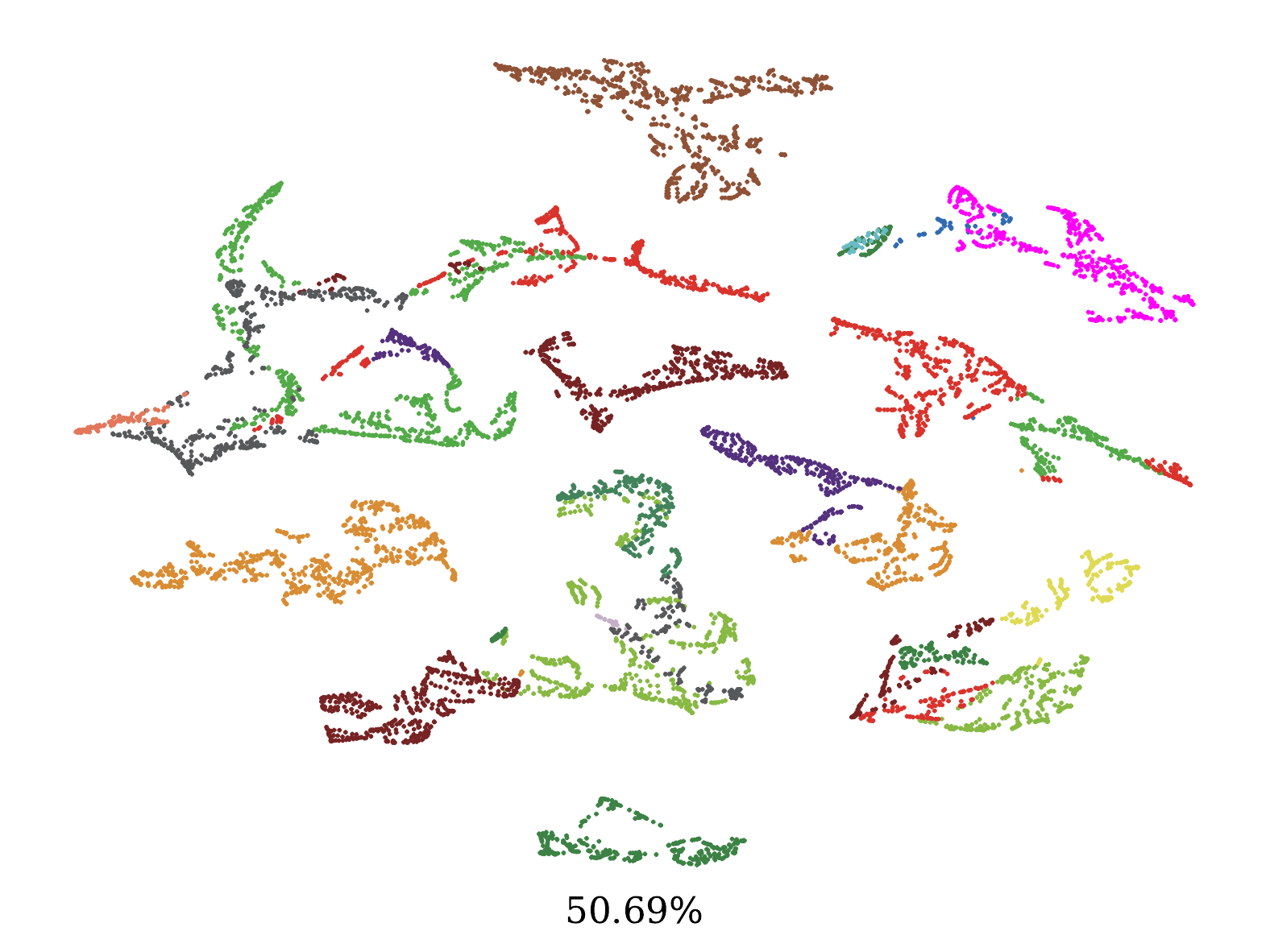}\includegraphics[width=0.4\columnwidth]{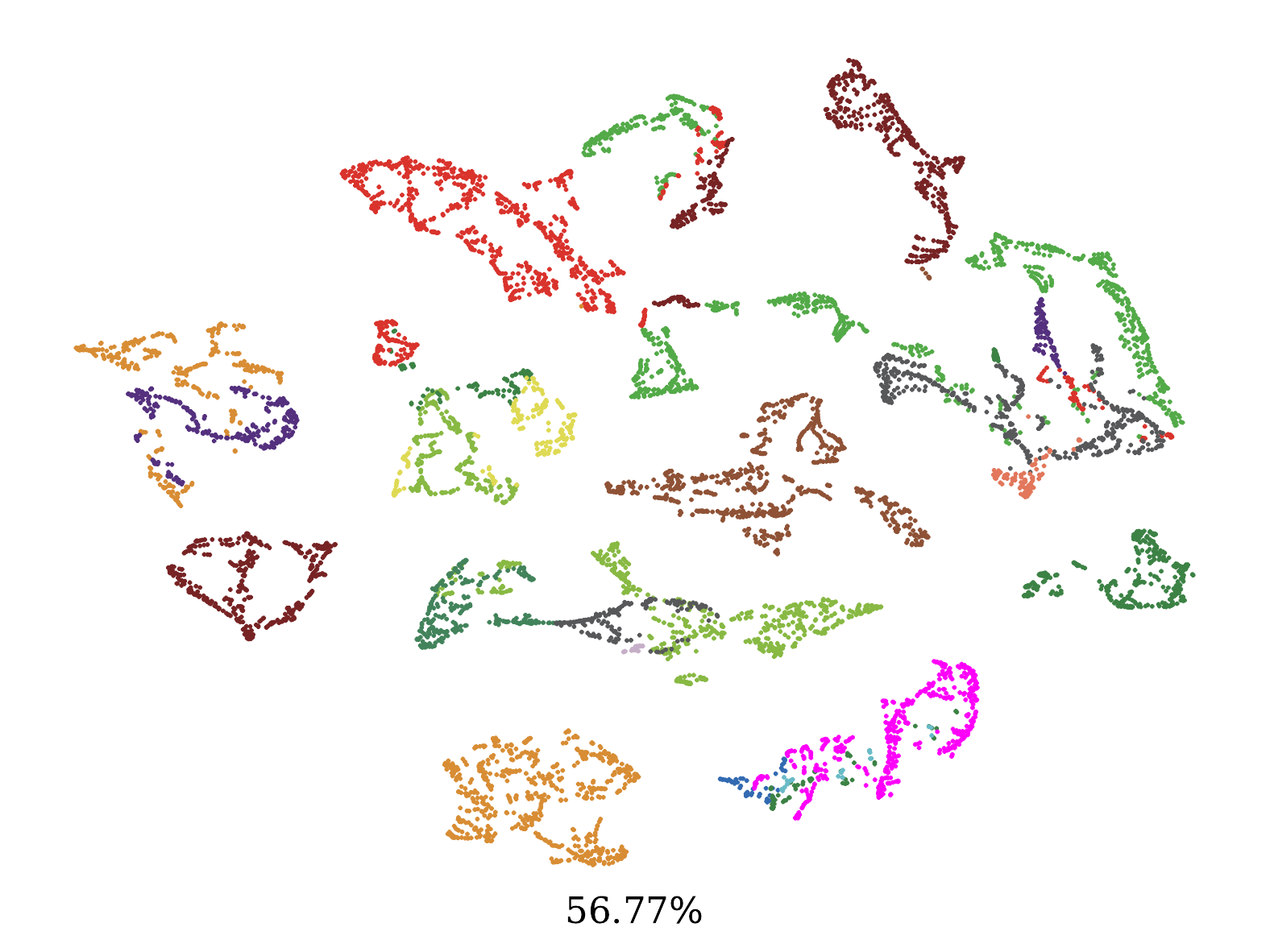}\includegraphics[width=0.4\columnwidth]{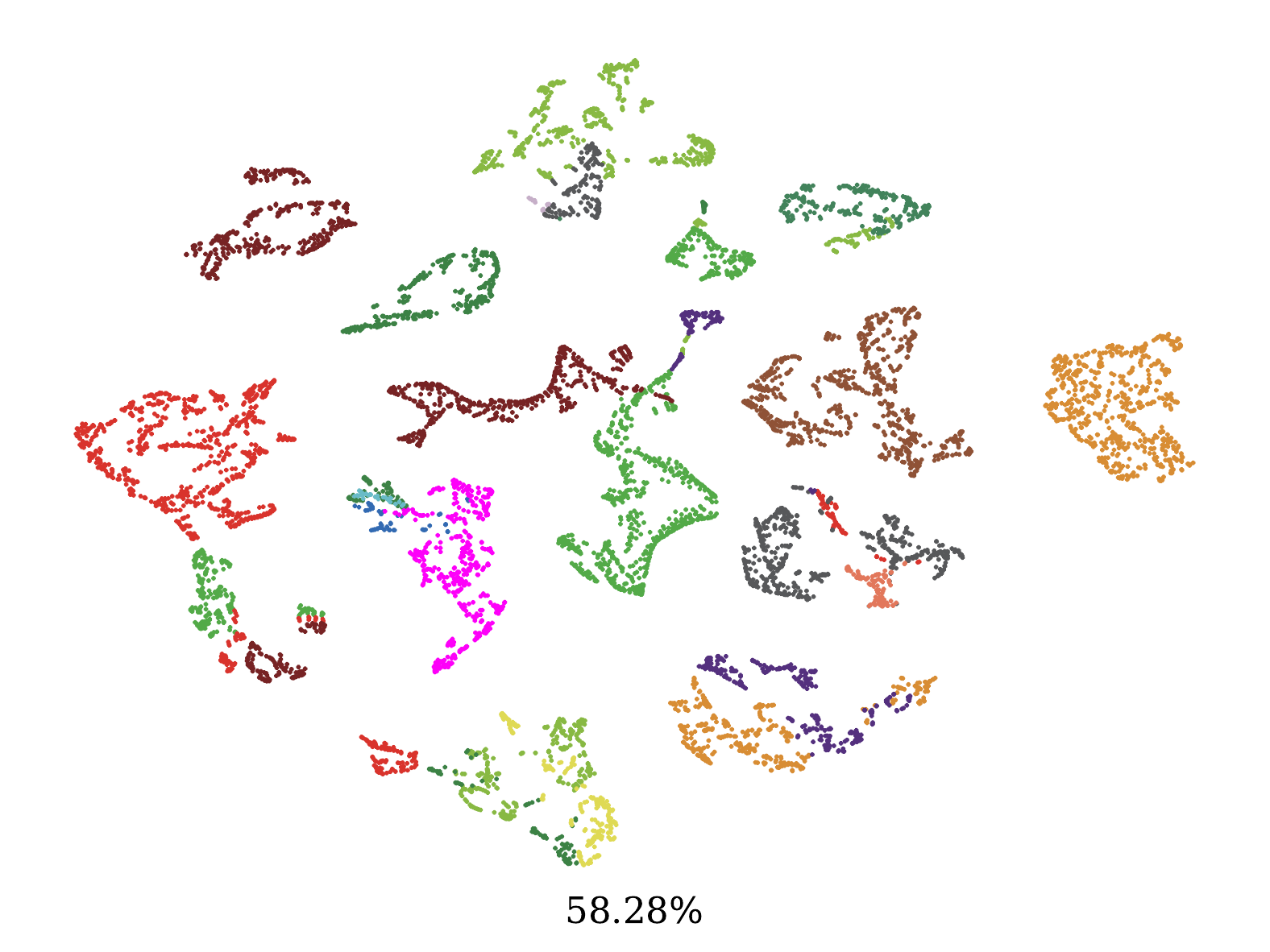}\includegraphics[width=0.4\columnwidth]{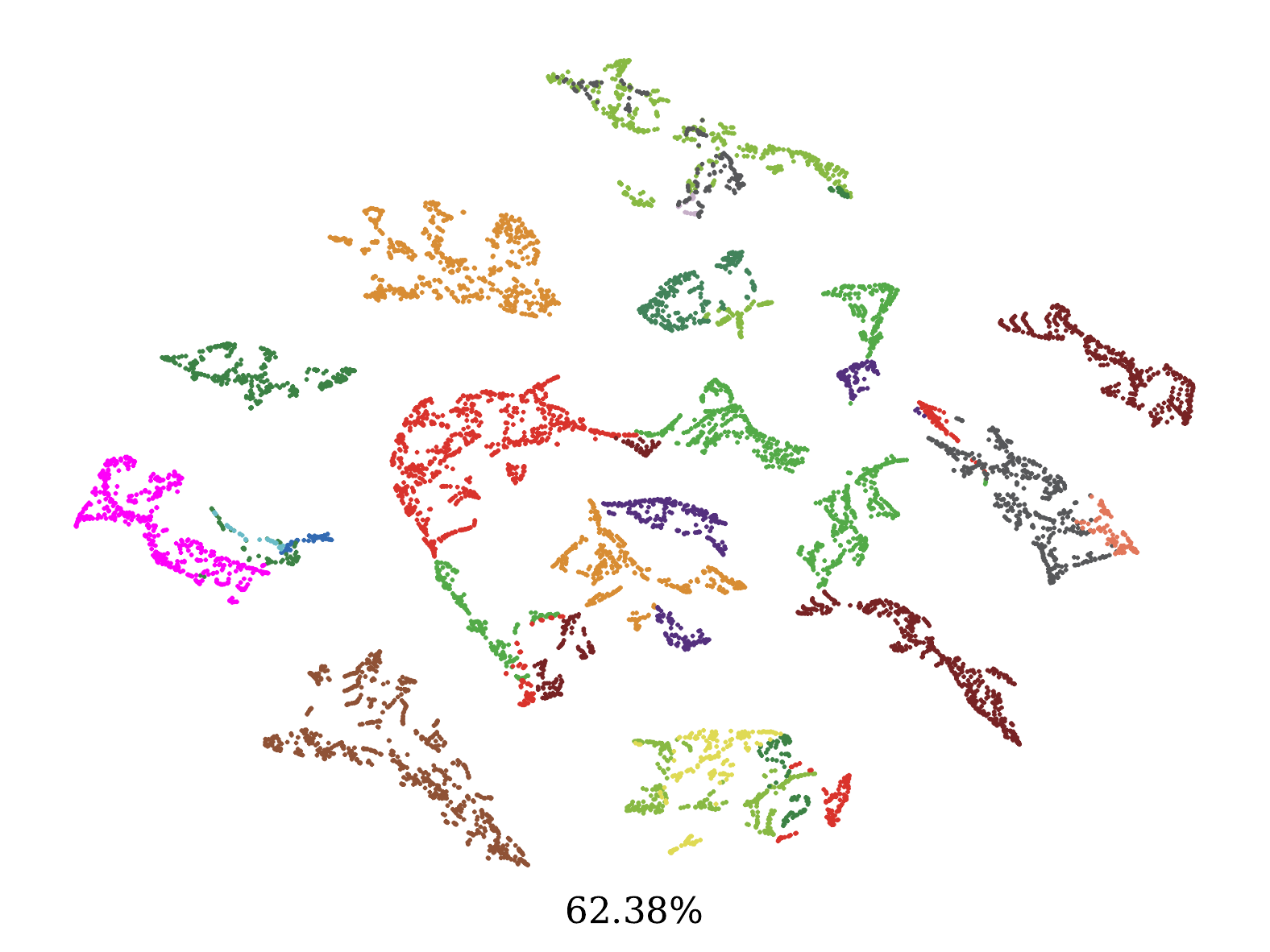}
	}
	
	\resizebox{2.\columnwidth}{!}{
		\includegraphics[width=0.4\columnwidth]{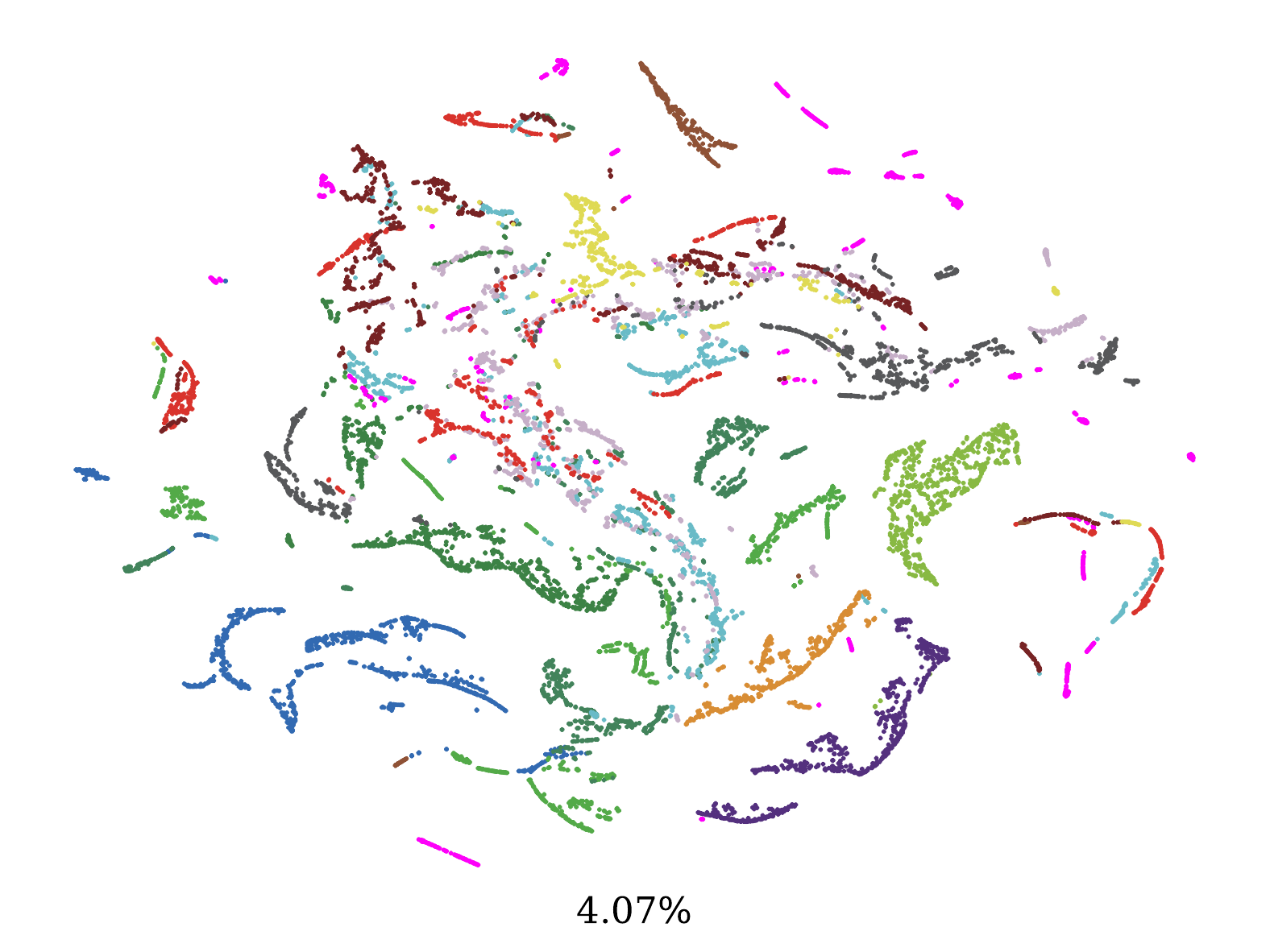}\includegraphics[width=0.4\columnwidth]{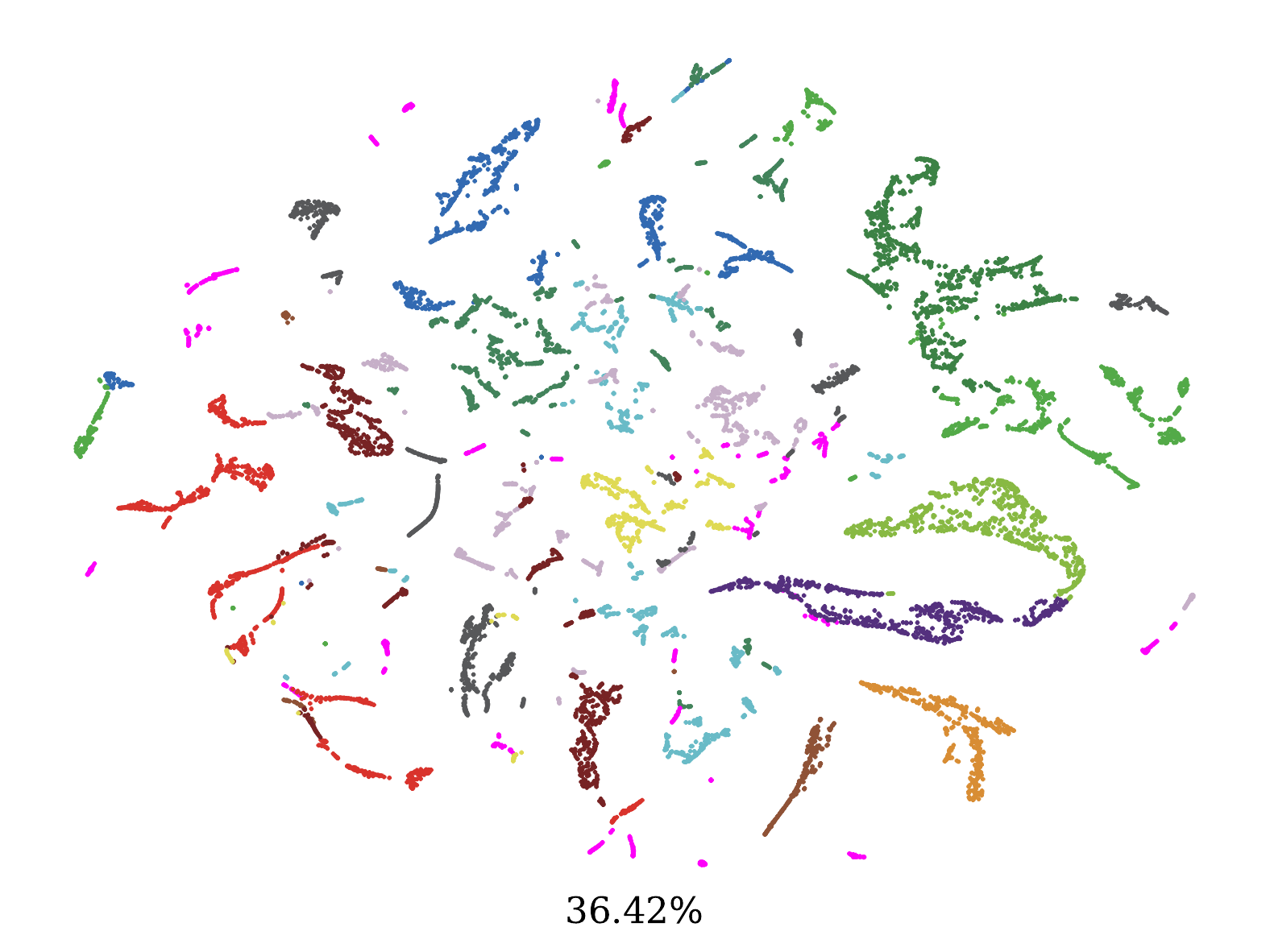}\includegraphics[width=0.4\columnwidth]{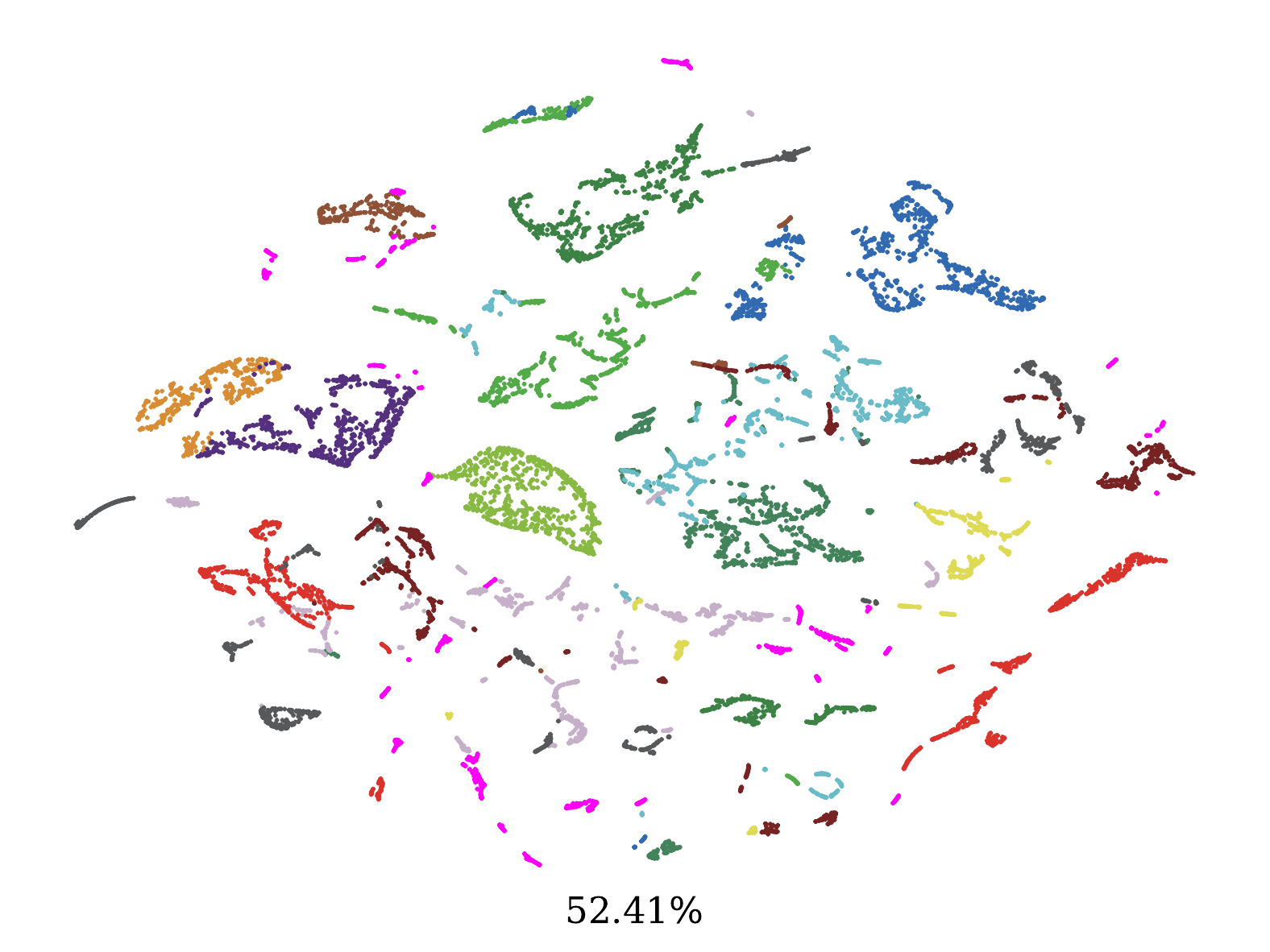}\includegraphics[width=0.4\columnwidth]{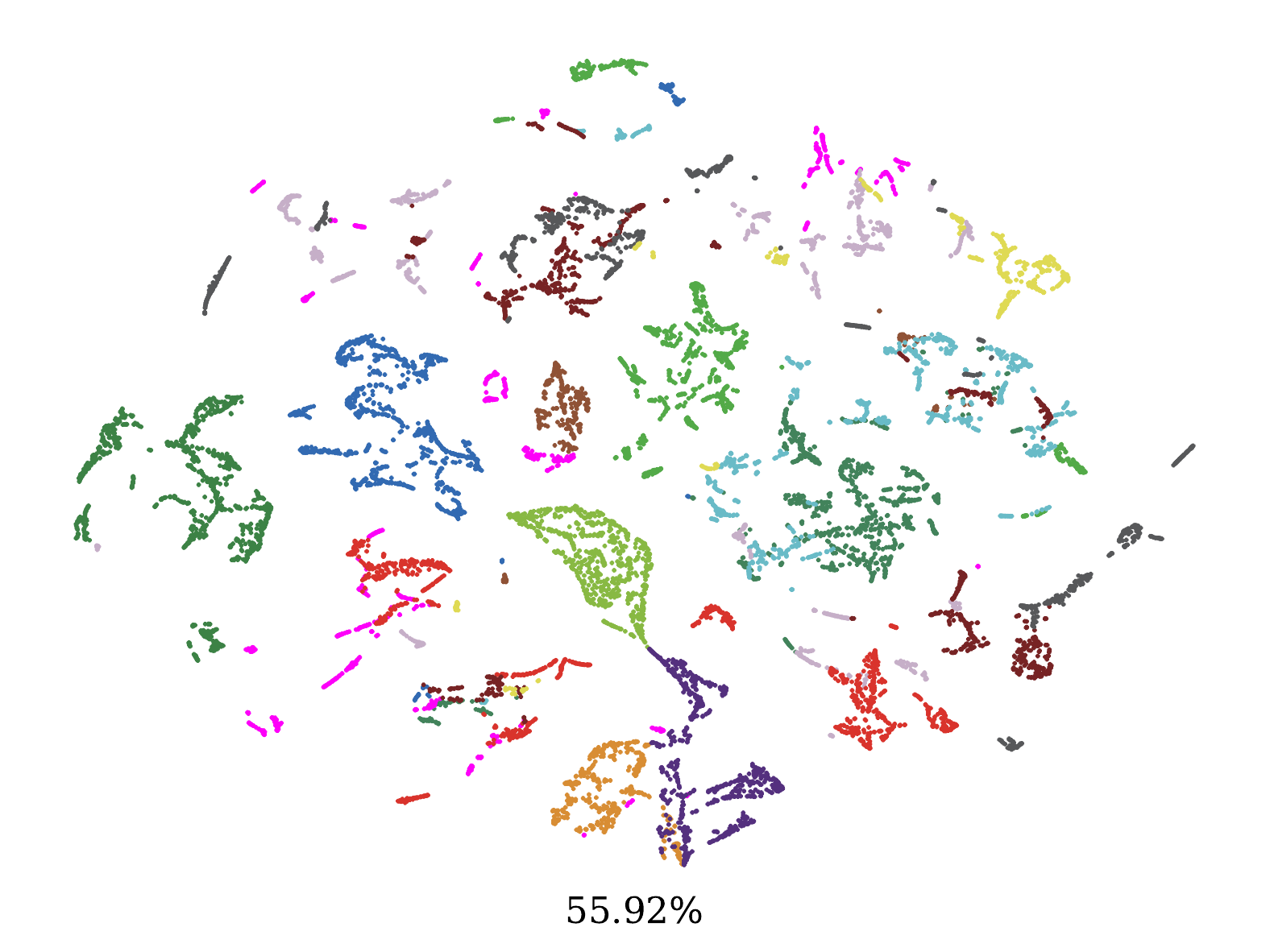}\includegraphics[width=0.4\columnwidth]{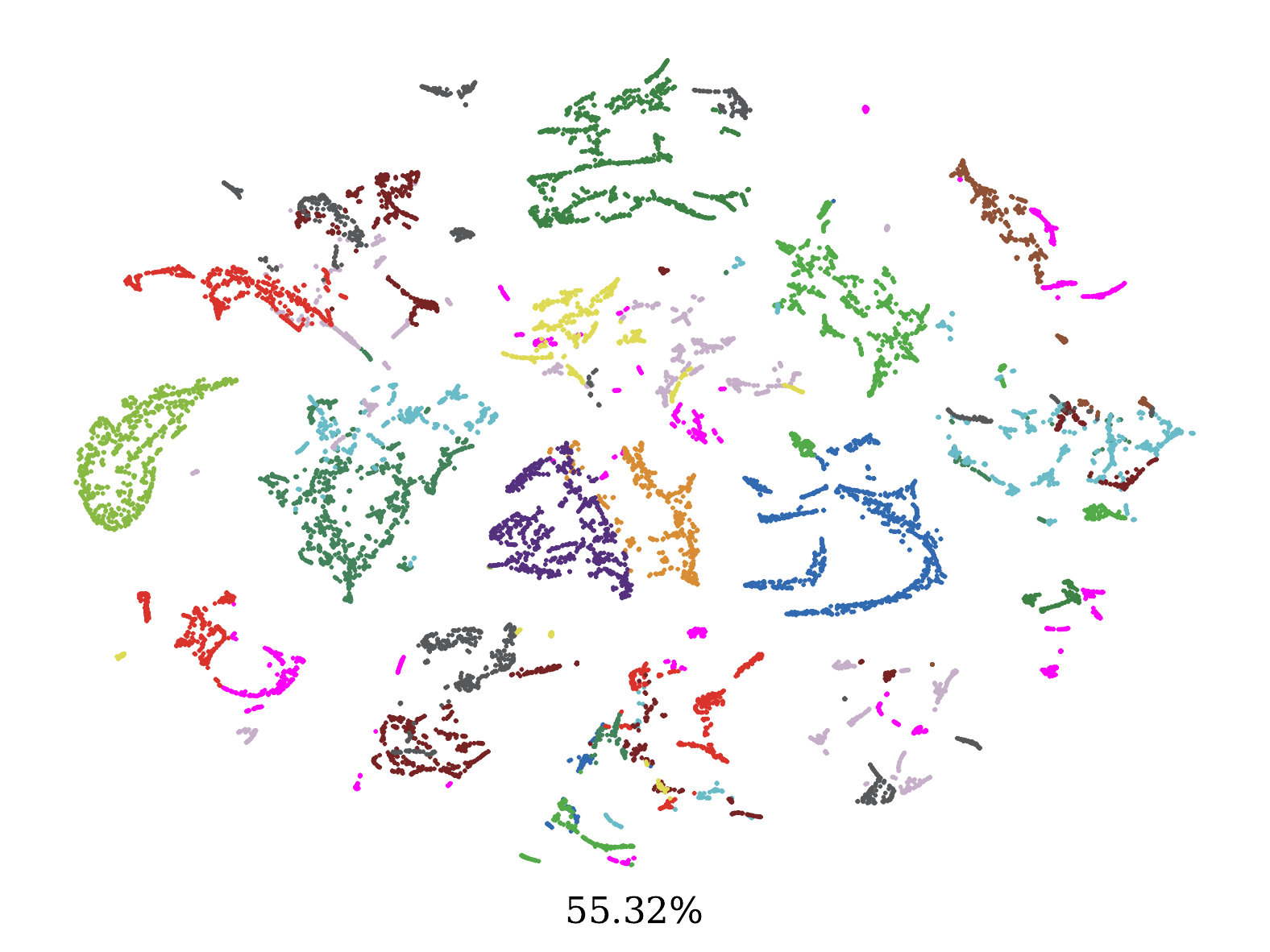}
	}
	
	\resizebox{2.\columnwidth}{!}{
		\includegraphics[width=0.4\columnwidth]{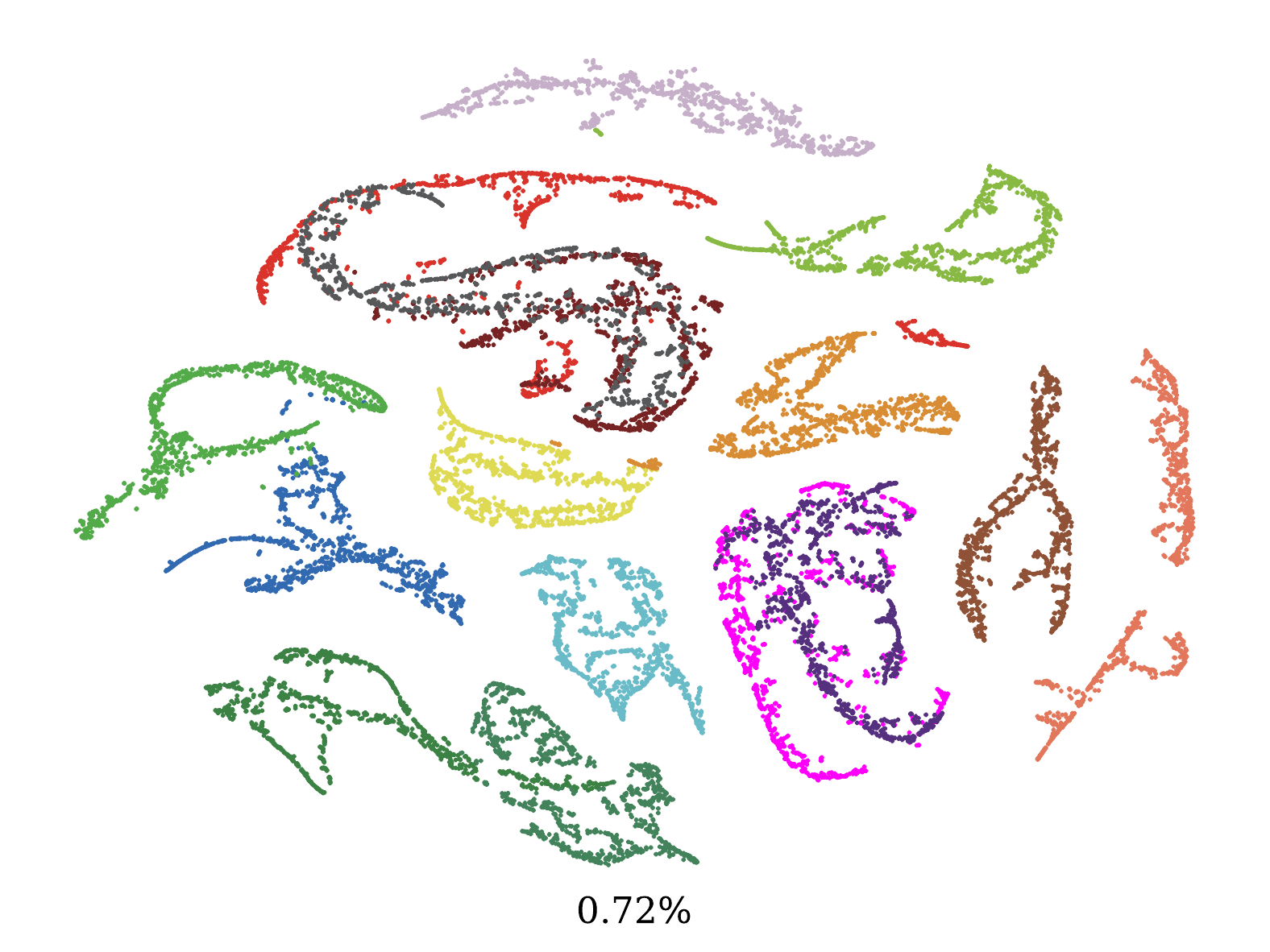}\includegraphics[width=0.4\columnwidth]{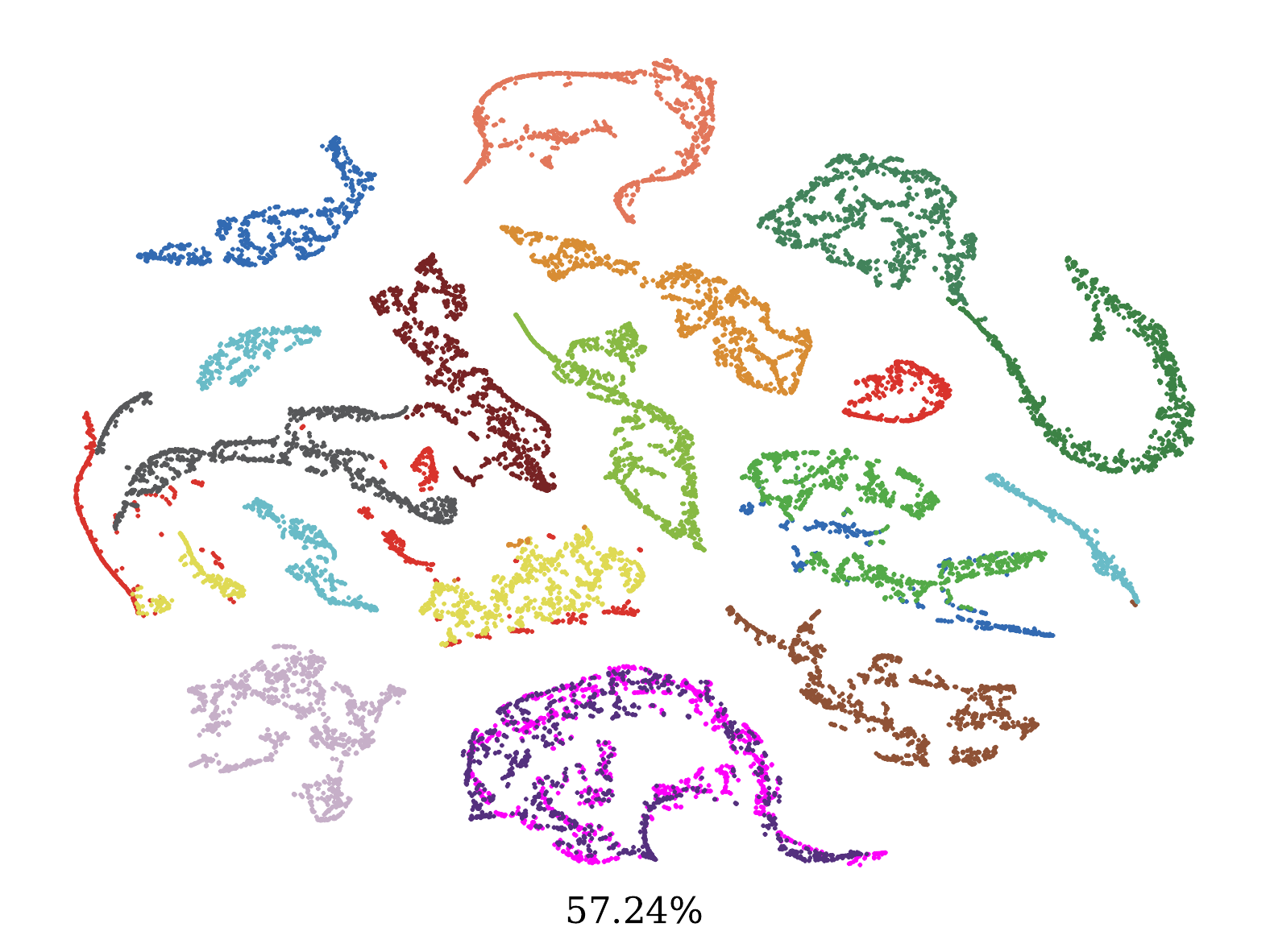}\includegraphics[width=0.4\columnwidth]{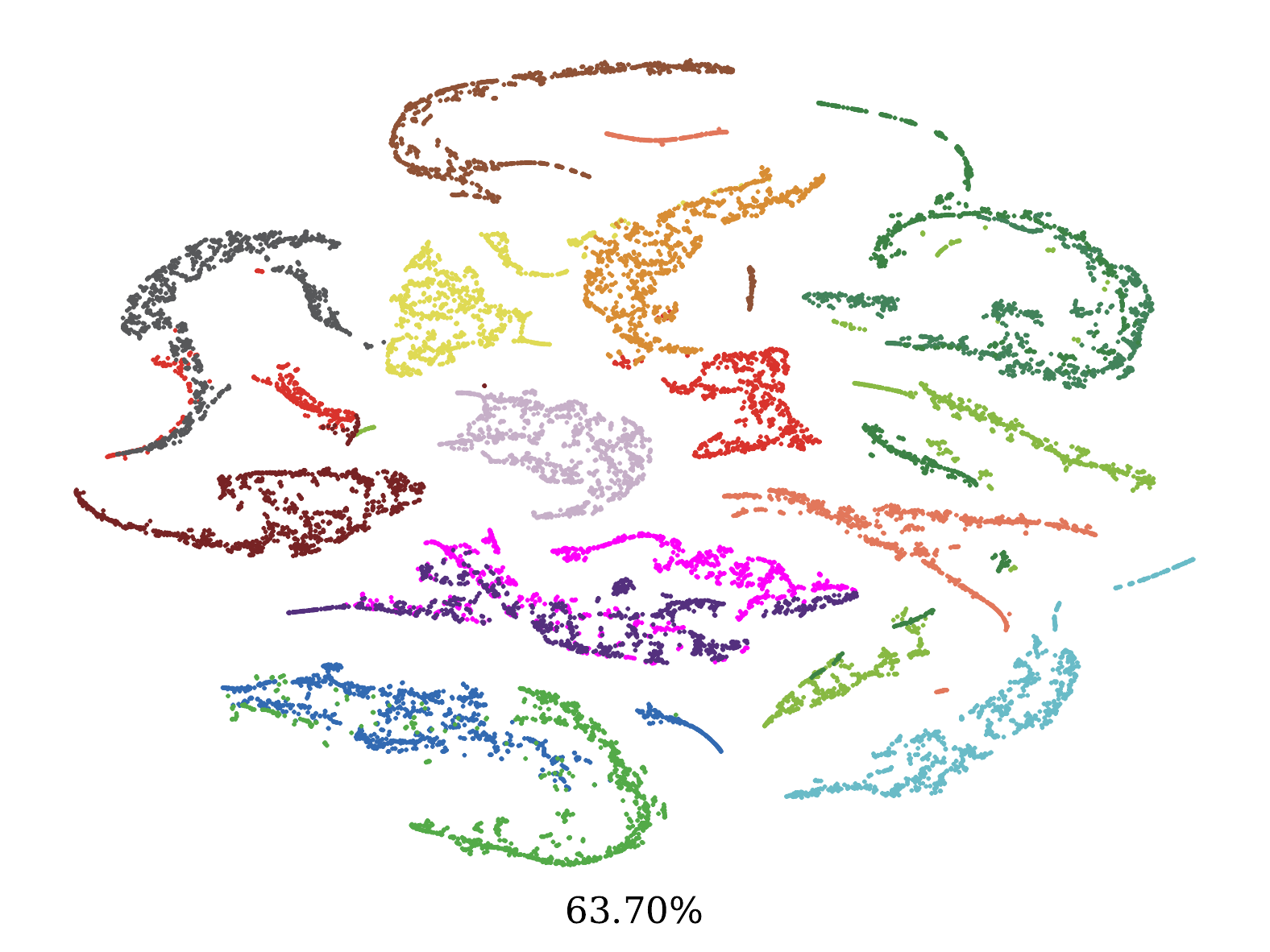}\includegraphics[width=0.4\columnwidth]{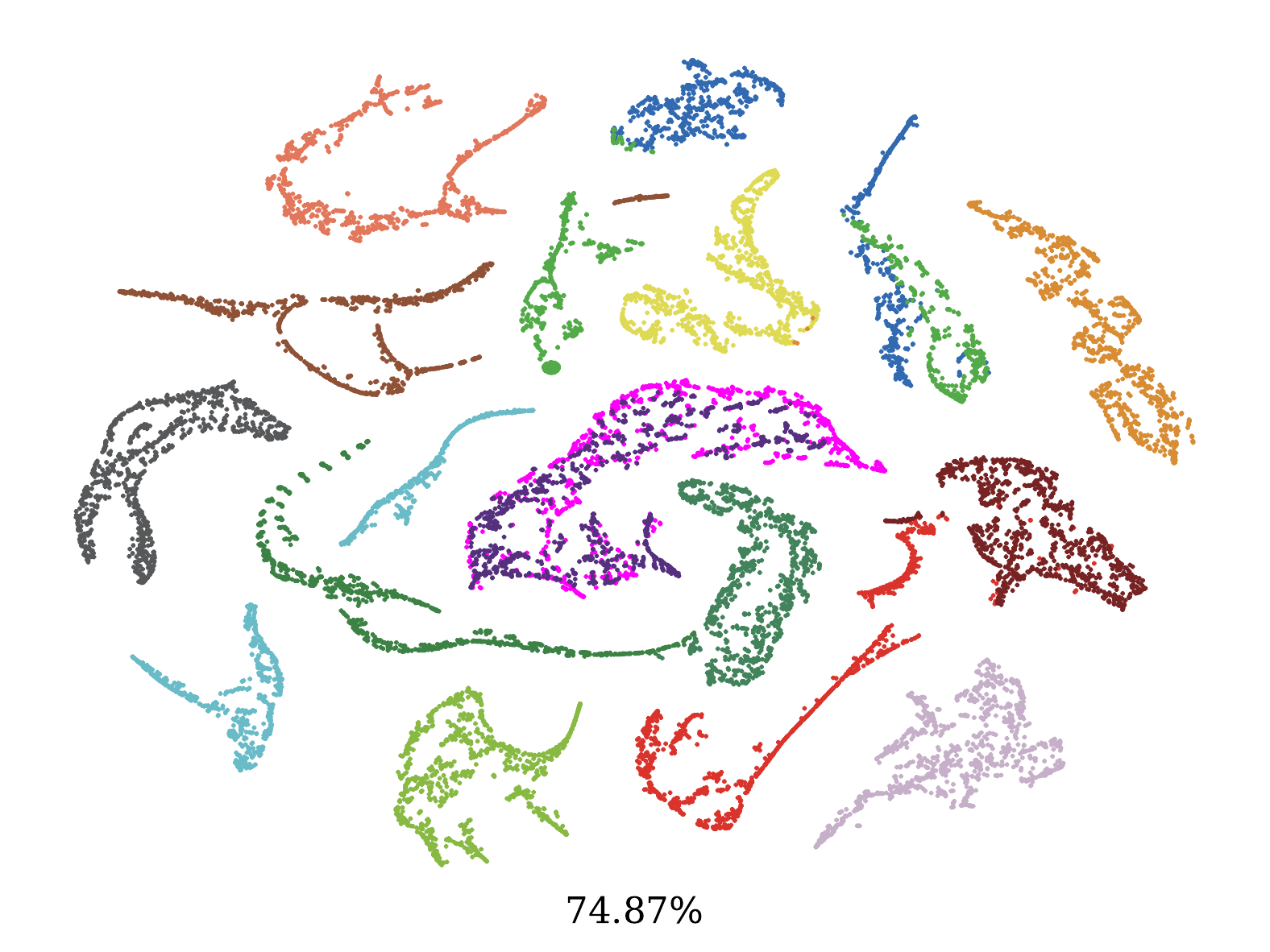}\includegraphics[width=0.4\columnwidth]{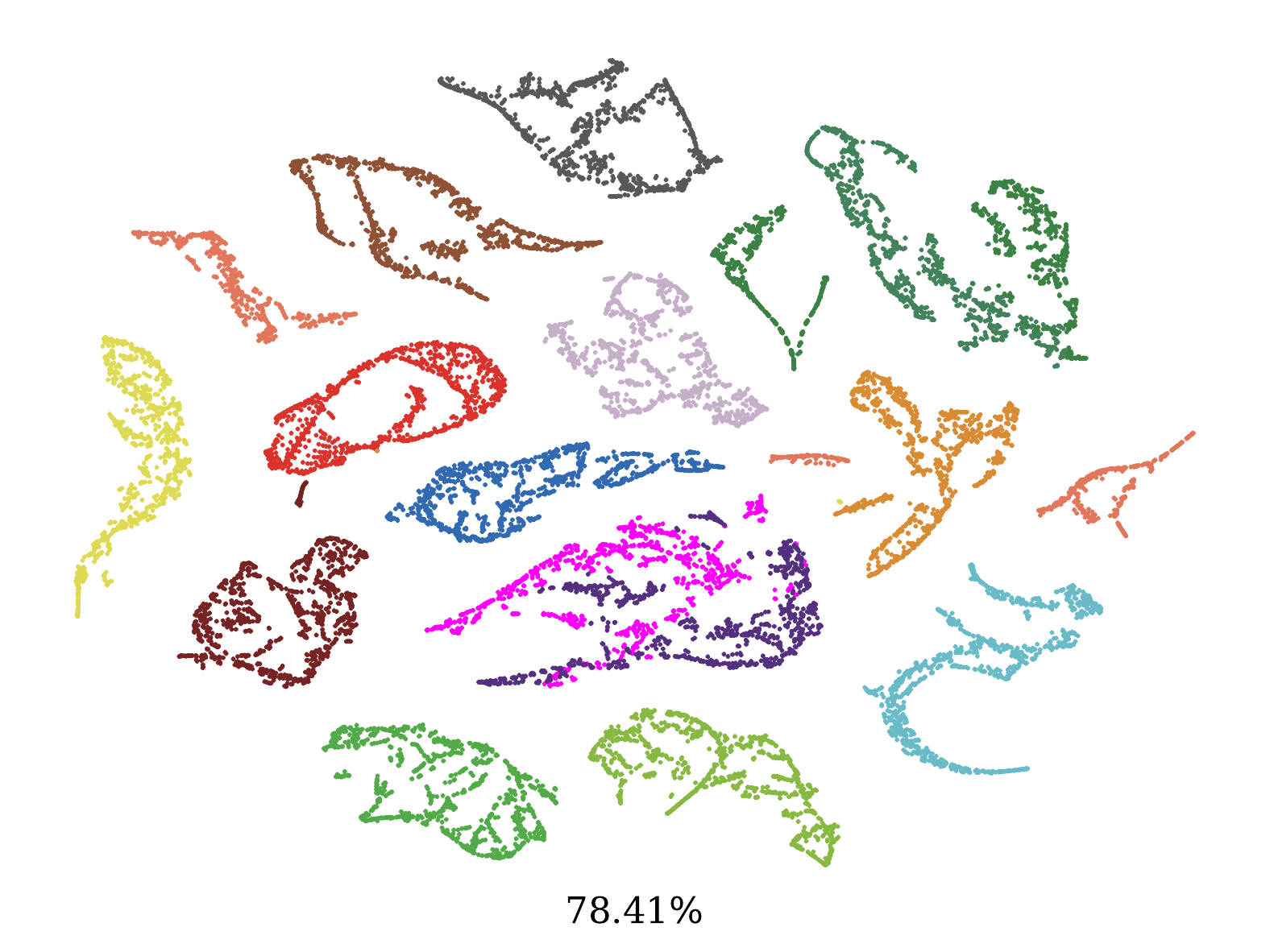}
	}
	
	\caption{t-SNE visualization on the Indian Pines (the first row), Houston (the
		second row), and Salinas (third row) datasets with an increasing training
		iteration. From left to right, the feature representations are obtained
		at epoch 0, 20, 40, 60, and 100, orderly. Each color indicates a cluster
		and the clustering ACC value is displayed under each visualization. \label{fig:t-sne}}
\end{figure*}

\begin{figure*}[tbh]
	\centering
	\resizebox{1.8\columnwidth}{!}{
		\begin{centering}
			\includegraphics[width=1.8\columnwidth]{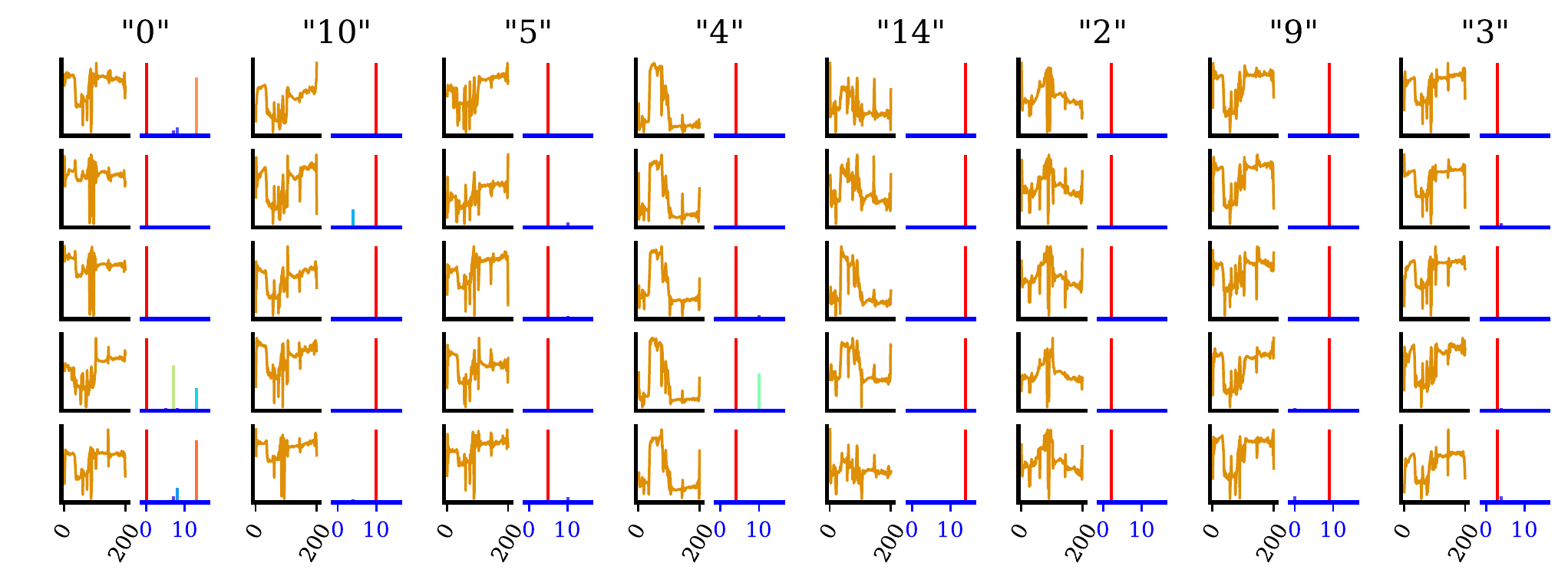}
			\par\end{centering}}
	\resizebox{1.8\columnwidth}{!}{
		\begin{centering}
			\includegraphics[width=1.8\columnwidth]{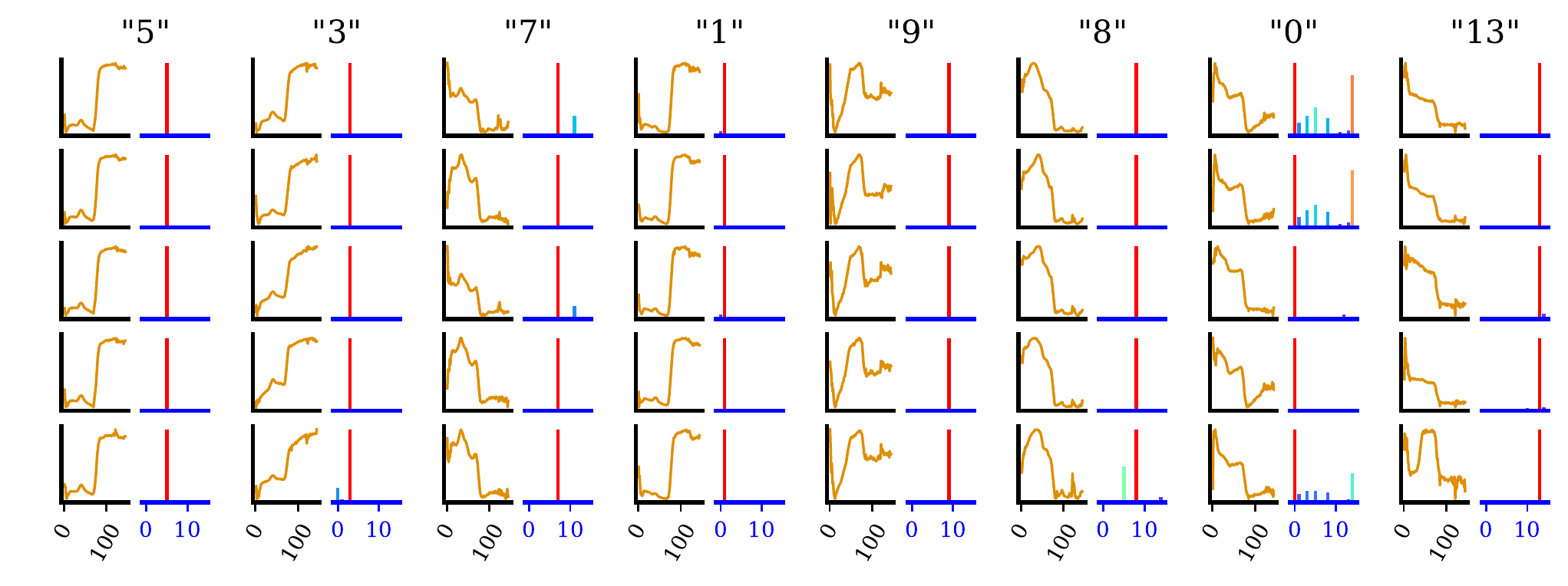}
			\par\end{centering}}
	\resizebox{1.8\columnwidth}{!}{
		\begin{centering}
			\includegraphics[width=1.8\columnwidth]{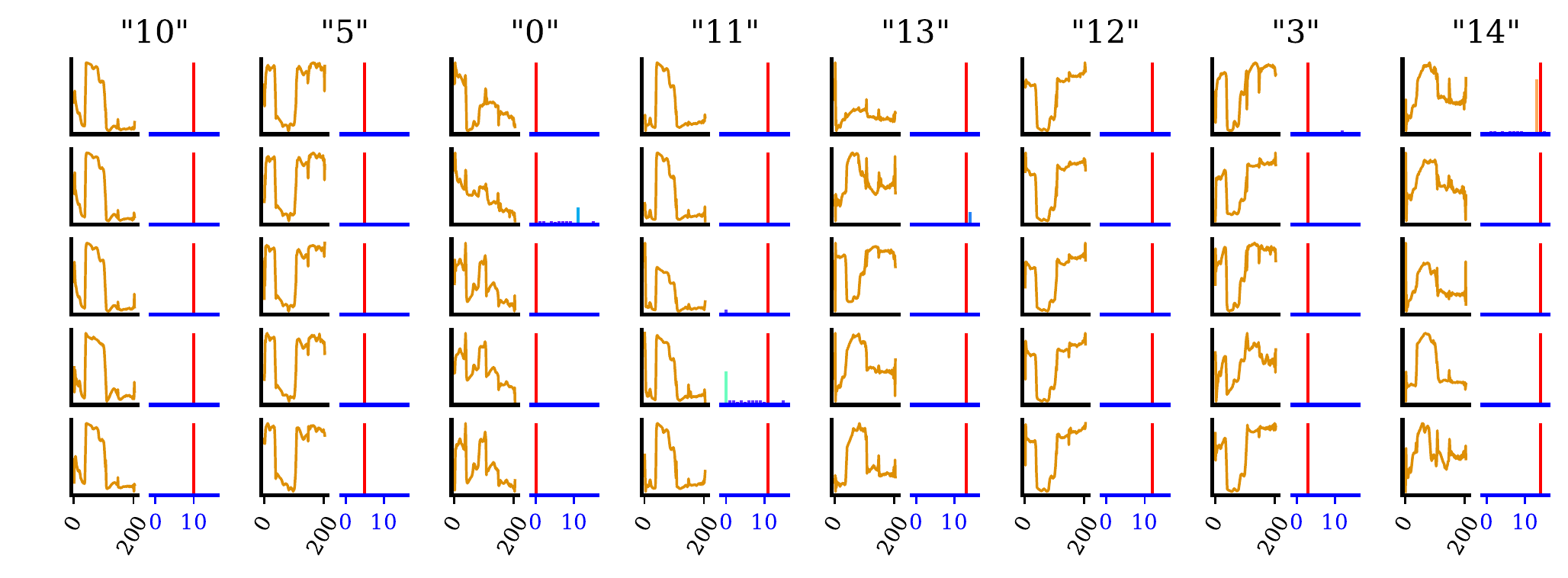}
			\par\end{centering}}
	\caption{The label representations of the first eight clusters on Indian Pines,
		Houston, and Salinas datasets. For each dataset, ground-truth labels
		are written on the upward side, the central spectrum curve of input
		patches are shown on the left side and corresponding predicted confidences
		are shown as histograms on the right side. We randomly select $5$
		cases from each class. \label{fig:label-visu}}
\end{figure*}

\subsection{Ablation Study and Parameter Analysis}

\subsubsection{Effect of Components of Objective}

To observe the effect of different components in the proposed objective
function, we conduct ablation studies on the three datasets by using
different combinations of objective terms. Specifically, based on
the InfoNCE loss term $\mathcal{L}_{\mathcal{W}}$ and the Barlow
Twins loss term $\mathcal{L}_{\mathcal{B}}$, we obtain five loss
function combinations: $\mathcal{L}_{\mathcal{W}}$ only, $\mathcal{L}_{\mathcal{B}}$
only, $\mathcal{L}_{\mathcal{W}}+\mathcal{L}_{\mathcal{W}}$, $\mathcal{L}_{\mathcal{B}}+\mathcal{L}_{\mathcal{B}}$
and $\mathcal{L}_{\mathcal{W}}+\mathcal{L}_{\mathcal{B}}$. The first
two combinations indicate we only consider either within-cluster loss
or between-cluster loss, while the third and the fourth ones signify
that both within- and between-cluster loss are implemented using the
same formulas as CC \cite{SSL-CC-Li-AAAI-21}. The results are given
in Table \ref{tab:ablation-loss}. It can be seen that our proposed
loss (i.e., $\mathcal{L}_{\mathcal{W}}+\mathcal{L}_{\mathcal{B}}$)
achieves more performance gain than others. Nevertheless, single loss
terms also show comparable performance by adjusting themselves parameters,
demonstrating the discriminative power of contrastive self-supervised
learning. Furthermore, the parameters $\tau$ and $\lambda$ involved
in $\mathcal{L}_{\mathcal{W}}$ and $\mathcal{L}_{\mathcal{B}}$ showcase
different behavior although they have a similar tendency. Following the conclusions
in \cite{ContrastLear-Wang-CVPR-21} and \cite{SSL-BarlowTwins-Jure-ICML-21},
smaller $\tau$ encourages the model to punish more on hard negative
samples, while smaller $\lambda$ tends to maintain the semantic invariability
between two views of inputs. The SSCC trained with $\mathcal{L}_{\mathcal{W}}+\mathcal{L}_{\mathcal{W}}$
and $\mathcal{L}_{\mathcal{B}}+\mathcal{L}_{\mathcal{B}}$ cannot
achieve significant improvement, even degenerating in the case of
$\mathcal{L}_{\mathcal{W}}+\mathcal{L}_{\mathcal{W}}$. It may be
because the InfoNCE highly relies on negative samples. In some certain
situations, for example, many false-negative samples in a mini-batch,
training with $\mathcal{L}_{\mathcal{W}}+\mathcal{L}_{\mathcal{W}}$
would lead to collapsed and unstable representations. That is why
$\mathcal{L}_{\mathcal{W}}$ often needs a large batch size for optimizing.

\subsubsection{Effect of Batch Size}

In order to explore the impact of batch size, we vary it in $\left\{ 64,128,256,512\right\} $
on three datasets. Table \ref{tab:ablation-batchsize} shows that
larger batch sizes consistently achieve better clustering accuracy
and better robustness than smaller ones. Our SSCC model typically
belongs to contrastive learning, thus the tendency is consistent with
the common observation on such kinds of learning models. As discussed
above, a significant reason is that the $\mathcal{L}_{\mathcal{W}}$
objective term requires enough negative pairs in a mini-batch to prevent
collapsed representation caused by the false-negative samples. Nevertheless,
the usage of Barlow Twins loss alleviates the strong dependency on
large batch size by redundancy reduction. We incorporate these two
losses together so that they can balance each other in the representation
space.


\subsubsection{Importance of Data Augmentation}

To explore the importance of augmentations, we perform SSCC on the
Indian Pines dataset under varying augmentation compositions of RandomCrop,
Resize, RandomFlip, RandomRotate, GaussianBlur, ErasurePixel, ErasureBand,
and PermuteBand. To better exhibit the influence, we apply only one
augmentation operation to each network branch. From Fig. \ref{fig:aug},
we observe that a single augmentation operation (diagonal entries)
is insufficient to guarantee robust clustering accuracy. In contrast,
the composition of two augmentation operations shows greater potential
in terms of accuracy. In particular, we observe that RandomCrop is
more effective than other spatial transformations. Although it increases
the difficulty of the contrast task, the spectral transformation has
a risk of destroying semantics and leading to the failure of capturing
spectral information. As a result, we conduct the spectral transformation
with a very small probability. Furthermore, the composition of multiple
augmentation operations significantly increases the diversity of data
distribution and thus is beneficial to the clustering robustness and
accuracy. 

\subsubsection{Effect of Patch Size}

To indicate the effect of input patch sizes, we evaluate the SSCC
model by varying input size from $5\times5$ to $21\times21$ with
an interval of $2\times2$. Thanks to the architecture of convolutions
followed by a global pooling, we can evaluate SSCC using arbitrary
patch size without modifying the model. We can observe from Fig. \ref{fig:patch}
that a suitable patch size is necessary for SSCC to achieve better
performance. Generally, HSI patches provide complementary information
from views of spectral and spatial, as well as associating the local
neighborhood relationship. However, due to the presence of mixed pixels
in HSI, there is an increased risk of introducing noisy pixels into
input patches when using a large patch size (e.g., $21\times21$). Furthermore,
the problem tends to get worse for HSIs with lower spatial resolution.
Consequently, from the results, the patch size should be empirically
set to no larger than $13\times13$. 

\subsection{Visualization}

\subsubsection{Convergence Analysis }

To analyze the convergence of the SSCC model, Fig. \ref{fig:loss-acc}
(a)-(c) visualize four clustering metrics against loss values for
$200$ training epochs. We observe that clustering metrics are remarkably
increased from a relatively low initial value to a stable and superior
value, while the loss value tends to be convergent after the $50$-th
epoch. This demonstrates that SSCC can gradually capture the spectral-spatial
clues by minimizing our defined within- and between-cluster criterion. 

To better understand the learning process, we quantify the discriminative
power of the learned label representations by defining the following
divergence value $S$, i.e.,
\begin{equation}
S=\frac{\sum_{c=1}^{C}\sum_{i\in\mathcal{N}_{c}}s\left(\boldsymbol{y}_{i},\bar{\boldsymbol{y}}_{c}\right)/\left|\mathcal{N}_{c}\right|}{\sum_{c=1}^{C}s\left(\bar{\boldsymbol{y}}_{c},\bar{\boldsymbol{y}}\right)},
\end{equation}
where $\bar{\boldsymbol{y}}_{c}$ and $\bar{\boldsymbol{y}}$ indicate
the mean vector of the $c$-th class and all samples, respectively,
and $\mathcal{N}_{c}$ denotes the index set of the $c$-th class.
Following Fisher's discriminant analysis \cite{LDA-1999}, the numerator
and denominator can be treated as within-cluster and between-cluster
scatter, respectively. Thus, a larger divergence value means that
better separability of the label representation is achieved. In Fig.
\ref{fig:loss-acc} (d), we show the divergence values by scaling
them into $\left[0,1\right]$. As the training epoch increases, the
divergence value of label representations is significantly increased,
signifying that SSCC's discriminant ability is enhanced obviously.
This also proves why our model can achieve remarkable results for
the HSI clustering task.

\subsubsection{Latent Representation Visualization}

Fig. \ref{fig:t-sne} shows the latent representations (i.e., outputs
of the backbone $f_{\theta}\left(\cdot\right)$) for Indian Pines,
Houston, and Salinas dataset. In this experiment, we use the t-SNE
\cite{t-SNE-JMLR-08} approach to reduce the dimensionality of latent
representations from initially $256$ into $2$. It can be seen that
the initial model can roughly separate samples into several limited
clusters. Such low-level discriminative ability results in low clustering
accuracy. Compared to the initial model, the well-trained model can
discover more underlying clusters with a higher clustering accuracy.
Furthermore, we can observe that the latent representations at epoch
100 exhibit obvious discriminative capability and uniformity. This
superiority is benefited from the mechanism of contrastive learning.

\subsubsection{Label Representation Visualization}

In Fig. \ref{fig:label-visu}, we qualitatively analyze the label
representations learned by SSCC on the three HSI datasets. For clarity,
we provide both the spectral curve and its corresponding prediction
confidence. We observe two obvious tendencies from the visualization.
First, the overall prediction confidence of all samples is quite sparse
though we didn't add any sparse regularization. Second, the neuron
that indicates the same cluster will be distinctly activated while
other neurons will be explicitly inhibited in label representations.
These two observations demonstrate that SSCC indeed attempts to capture
 high-level semantic differences, rather than a simple combination
of lower-level representations. In addition, we find that SSCC tends
to assign relatively larger confidence to those neurons whose corresponding
spectral curves are similar to the true ones, e.g., class 13 and class
14 on the Salinas. This is mainly caused by the spectral variability
of HSI. Furthermore, it also may be the main factors that lead to
the mis-clustering of those similar land-cover objects. 

\section{Conclusions\label{sec:Conclusions}}

We have presented a novel one-stage online SSCC clustering approach
for unsupervised learning of large-scale hyperspectral data. The proposed approach
follows a contrastive learning pipeline and consists of a spectral-spatial
augmentation pool, a backbone network aiming to extract high-level
representations and a clustering-specific projection head. We designed
a novel objective function based on dual contrastive objectives which
implicitly maximizes the within-cluster similarity and minimizes the between-cluster redundancy, simultaneously. The SSCC model is characterized by end-to-end training
with mini-batch samples and online clustering, making it scalable
to large-scale HSI and possible for being applicable in practice. Experimental
results on real HSI datasets show that SSCC can achieve the state-of-the-art
clustering performance with remarkable margins over previous works. 

The success of SSCC not only narrows the gap between unsupervised
methods and supervised methods but also offers a powerful alternative
for designing unsupervised methods in the remote sensing community.
Future work may include exploring more effective self-supervised schemes
to improve the discriminative power and the generalization capability of SSCC to other downstream
tasks. For the first problem, an increasing number of self-supervised
learning models have been devised recently and some helpful ideas
including hard negative mining are worth to be investigated. For the second
problem, it is interesting to transfer knowledge from SSCC to
other specific tasks, e.g., HSI segmentation and object detection. 

\section*{}

\bibliographystyle{IEEEtran}
\bibliography{references}

\end{document}